\documentclass[nohyperref]{article}

\usepackage{microtype}
\usepackage{graphicx}
\usepackage{subfigure}
\usepackage{booktabs} % for professional tables
\usepackage{stmaryrd}
\usepackage{dsfont}
\usepackage{mathtools, nccmath}

\newcommand{\rvline}{\hspace*{-\arraycolsep}\vline\hspace*{-\arraycolsep}}

\usepackage[accepted]{icml2022}

\usepackage{amsmath}
\usepackage{amssymb}
\usepackage{mathtools}
\usepackage{amsthm}

\usepackage[capitalize,noabbrev]{cleveref}

\theoremstyle{plain}
\newtheorem{theorem}{Theorem}[section]

\newtheorem{lemma}[theorem]{Lemma}

\theoremstyle{definition}

\theoremstyle{remark}

\usepackage[textsize=tiny]{todonotes}

\icmltitlerunning{Submission and Formatting Instructions for ICML 2022}

\begin{document}

\twocolumn[
\icmltitle{On the Initialisation of Wide Low-Rank Feedforward Neural Networks \\}

% It is OKAY to include author information, even for blind
% submissions: the style file will automatically remove it for you
% unless you've provided the [accepted] option to the icml2022
% package.

% List of affiliations: The first argument should be a (short)
% identifier you will use later to specify author affiliations
% Academic affiliations should list Department, University, City, Region, Country
% Industry affiliations should list Company, City, Region, Country

% You can specify symbols, otherwise they are numbered in order.
% Ideally, you should not use this facility. Affiliations will be numbered
% in order of appearance and this is the preferred way.
%\icmlsetsymbol{equal}{*}

\begin{icmlauthorlist}
\icmlauthor{Thiziri Nait Saada}{maths}
\icmlauthor{Jared Tanner}{maths,turing}
\end{icmlauthorlist}

\icmlaffiliation{maths}{Mathematical Institute, University of Oxford, UK}
\icmlaffiliation{turing}{The Alan Turing Institute, London, UK}

\icmlcorrespondingauthor{Thiziri Nait Saada}{thiziri.naitsaada@maths.ox.ac.uk}

% You may provide any keywords that you
% find helpful for describing your paper; these are used to populate
% the "keywords" metadata in the PDF but will not be shown in the document
\icmlkeywords{Random Networks, Initialisation, Edge-of-chaos, Low-rank, Deep Learning}

\vskip 0.3in
]

% this must go after the closing bracket ] following \twocolumn[ ...

% This command actually creates the footnote in the first column
% listing the affiliations and the copyright notice.
% The command takes one argument, which is text to display at the start of the footnote.
% The \icmlEqualContribution command is standard text for equal contribution.
% Remove it (just {}) if you do not need this facility.

%\printAffiliationsAndNotice{}  % leave blank if no need to mention equal contribution
\printAffiliationsAndNotice{} % otherwise use the standard text.

\begin{abstract}
The edge-of-chaos dynamics of wide randomly initialized low-rank feedforward networks are analyzed.  Formulae for the optimal weight and bias variances are extended from the full-rank to low-rank setting and are shown to follow from multiplicative scaling.  The principle second order effect, the variance of the input-output Jacobian, is derived and shown to increase as the rank to width ratio decreases.  These results inform practitioners how to randomly initialize feedforward networks with a reduced number of learnable parameters while in the same ambient dimension, allowing reductions in the computational cost and memory constraints of the associated network.
%Abstracts must be a single paragraph, ideally between 4--6 sentences long.
\end{abstract}

\section{Introduction}
\label{sec1:introduction}

% 0th paragraph on infinitely wide NN?
%Infinitely wide neural neural networks have recently drawn some attention for their simple mathematical representation as Gaussian processes \cite{neal_1995}, \cite{Matthews_2018}, \cite{Lee_2017} and their ability to describe better phenomena emerging in increasingly larger Deep Learning models designed with hundreds of hidden neurons at each layer. Before even training a model, it is important to reckon the importance of well initialising its parameters. Indeed, objective functions in Deep Learning are usually highly non convex and state of the art optimisation methods performance rely on the initial point. 
 
% First paragraph
%In the context of infinitely wide neural networks, the Edge of Chaos tells us how to initialise deep networks so that the geometrical information of the data is preserved when being propagated along the network's layers. In \cite{Poole_2016}, the authors define the two quantities at stake characterising the data geometry as being the length (in the sense of the $L_2$ norm) of the entries and the angle or correlation between inputs (in the sense of the canonical inner product). Alternatively, the authors in \cite{Tanner_2021} examine this initialisation scheme from an information theory perspective and note that the mutual information flow is better preserved on the Edge of Chaos. 

Neural networks being applied to new settings, limiting transfer learning, are typically initialized with i.i.d. random entries.  The edge-of-chaos theory of \cite{Poole_2016} determine the appropriate scaling of the weight matrices and biases so that intermediate layer representations \eqref{eq:h} and the median of the input-output Jacobian's spectra \eqref{eq:J} are to first order independent of the layer.  Without this normalization there is typically an {\em exponential} growth in the magnitude of these intermediate representations and gradients as they progress between layers of the network; such a disparity of scale inhibits the early training of the network \cite{Glorot_2010}.

For instance, consider an untrained fully connected neural network whose weights and biases are set to be respectively identically and independently distributed with respect to a Gaussian distributions: $W^{(l)}_{ij} \sim \mathcal{N}(0, \frac{\sigma^2_W}{N_{l-1}})$, $b^{(l)}_i \sim \mathcal{N}(0, \sigma^2_b)$ with $N_l$ the width at layer $l$.  Starting such a network, with nonlinear activation $\phi : \mathbb{R} \to \mathbb{R}$, from an input vector $z^0:=x^0 \in \mathbb{R}^{N_0}$, the data propagation is then given by the following equations, 
\begin{align}\label{eq:h}
    h_j^{(l)} = \sum\limits_{k=1}^{N_{l-1}} W^{(l)}_{jk} z^{(l)}_k + b^{(l)}_j, \qquad z^{(l)}_k = \phi(h^{(l-1)}_k)
\end{align}
where we call $h^{(l)}$ the preactivation vector at layer $l$.

It has been shown by \cite{Poole_2016} that the preactivation vectors $h^{l}$ have geometric properties of 
%there exists a mapping from a layer to the next one describing the forward propagation of some geometrical properties of the preactivation vectors, respectively the 
length $q^{l}:=N_l^{-1}\left(h^{l}\right)^Th^{l}$ and the pairwise covariance $q^{l}_{12}:=N_l^{-1}\left(h_1^{l}\right)^T h_2^{l}$ of two inputs $x^{0,1}$ and $x^{0,2}$ which propagate through the network according to functions of the network entries' variances and nonlinear activation $(\sigma_b, \sigma_W, \phi)$.  These propagation maps were computed by \cite{Poole_2016}  in the limiting setting of infinitely wide networks and either i.i.d. Gaussian entries or scaled randomly drawn orthnormal matrices.  Here we extend this setting to their low-rank analogous.

Consider rank $r_l := \gamma_l N_l$  weight matrices, $W^{(l)} \in \mathbb{R}^{N_{l}\times N_{l-1}}$, formed as
\begin{align}\label{definition_weights}
W^{(l)}_{ij} = \sum\limits_{k=1}^{r_l} \alpha_{k,j}^{(l)} (C_k^{l})_i,
\end{align}
where the scalars $\left(\alpha^{(l)}_{k,i}\right)_{1\leq i \leq N_{l-1}}\in \mathbb{R} \overset{\text{iid}}{\sim} \mathcal{N}(0,\frac{\sigma_{\alpha}^2}{N_{l-1}})$ and the columns $C^{(l)}_1, \dots, C^{(l)}_{r_l}$ are drawn jointly as the matrix $C^{(l)}:=[C^{(l)}_1, \dots, C^{(l)}_{r_l}]\in\mathbb{R}^{N_l\times r_l}$ from the Grassmannian of rank $r$ matrices with orthonormal columns having zero mean and variance $1/N_l$.
%\overset{\text{iid}}{\sim} \mathcal{N}(0, \frac{1}{N_l}) $. Additionally, the random direction vectors are normalised, i.e. for any $k$, $||C_{k}^{(l)}||_2 = 1$. 
Similarly, consider bias vectors within the same column span as $W^{(l)}$, given by $b^{(l)} (C^{(l)}_1 + \dots + C^{(l)}_{r_l})$, where $b^{(l)}\in \mathbb{R} \sim \mathcal{N}(0, \sigma_b^2)$.  It is shown in Appendix \ref{app:h_distribution} that, in the large width limit, the preactivation vector $h^{(l)}$ follows a Gaussian distribution over the $r-$dimensional column span of $W^{(l)}$ with a non-diagonal covariance; this differs from the full rank setting in \cite{Poole_2016} where the entries in \eqref{eq:h} are independent.

We extend the pre-activation length and correlation maps to this low-rank setting:
\begin{align}
q^l & = \gamma_l \bigg( \sigma^2_{\alpha} \int_{\mathbb{R}} \phi^2(\sqrt{q^{l-1}}z) Dz + \sigma_b^2\bigg) \label{eq:qmap}\\
&:= \mathcal{V}(q^{(l-1)}|\sigma_{\alpha}, \sigma_b, \gamma_l)
\end{align}
where $Dz:=\frac{1}{\sqrt{2\pi}} e^{-\frac{z^2}{2}}dz$ is the Gaussian probability measure, and 
\begin{align}
q^l_{12} &= \gamma_l \bigg( \sigma^2_{\alpha} \int_{\mathbb{R}^2} \phi(u_1)\phi(u_2) Dz_1 Dz_2 + \sigma^2_b \bigg),\label{eq:q12}\\ 
&:= \mathcal{C}(q_{ab}^{l-1}, q_{aa}^{l-1},  q_{bb}^{l-1} | \sigma_{\alpha}, \sigma_b,\gamma_{l}) \label{eq:Cmap}
\end{align}
with $u_1 =\sqrt{q^{l-1}_{11}}z_1, u_2 = \sqrt{q^{l-1}_{22}}(c_{12}^{l-1} z_1 + \sqrt{1-(c_{12}^{l-1})^2}z_2$ and 
\begin{align}c_{12}^{l} = q_{12}^l (q_{11}^l q_{22}^l)^{-\frac{1}{2}}.\label{eq:c}
\end{align}
Equations \eqref{eq:qmap} and \eqref{eq:q12} are derived in \cref{SM:length_recursion} and \cref{SM:covariance_recursion} respectively.  
%If we rescale such that $\sigma^2_{\alpha} = \frac{\sigma^2_{W}}{\gamma}$, we recover the result from \cite{Poole_2016}.
These equations exactly recover the equations by \cite{Poole_2016} when $\gamma_l=1$, and show that by appropriately rescaling $\sigma_2^2$ and $\sigma_b^2$ by $\gamma_r$ the low-rank maps remain consistent with the full rank setting.

%satisfy the following mappings, in the large width $N_l$ limit, from one layer to the next:
%\begin{align}
%q^l  & = \sigma_W^2 \int_{\mathbb{R}} \phi(\sqrt{q^{l-1}} z)^2 Dz + \sigma_b^2 \nonumber \\
%& := \mathcal{V}(q^{l-1}|\sigma_W, \sigma_b) \label{eq:Vmap} \\
%q^{l}_{ab} & = \sigma_W^2 \int_{\mathbb{R}^2} \phi(u_1)  \phi(u_2) Dz_1 Dz_2 + \sigma_b^2 \nonumber \\
%& := \mathcal{C}(q_{ab}^{l-1}, q_{aa}^{l-1},  q_{bb}^{l-1} | \sigma_W, \sigma_b) \label{eq:Cmap}
%\end{align}
%$u_1 = \sqrt{q_{aa}^{l-1}} z_1, u_2 = \sqrt{q_{bb}^{l-1}} \big[ c^{l-1}_{ab} z_1 + \sqrt{1 -(c^{l-1}_{ab})^2}z_2 \big]$ and the correlation is given by $ c^l_{ab} = q^l_{ab}(q^l_{aa} q^l_{bb})^{-\frac{1}{2}}$.

These two mappings \eqref{eq:qmap} and \eqref{eq:q12} are functions of the network entries variances, the rank at each layer $\gamma_l$ and the nonlinear activation $(\sigma_b, \sigma_W, \gamma_l, \phi)$ which determine the existence of eventual stable fixed points of $q^{l}$ and $q^{l}_{ab}$ as well as the dynamics they follow through the network.

The dominant quantity determining the dynamics of the network is
\begin{align}\label{eq:chi}
\chi_{\gamma} &:= \gamma \sigma^2_{\alpha} \int_{\mathbb{R}} \bigg(\phi^\prime(\sqrt{q^*}z)\bigg)^2 Dz
\end{align}
which is equal to two fundamental quantities.  First, $\chi_{\gamma}$ is equal to the gradient of the correlation function \eqref{eq:c} evaluated at correlation $c_{12}^l=1$, 
\begin{align}\label{eq:Cprime}
\chi_{\gamma} = \frac{\partial c^l_{12}}{\partial c^{l-1}_{12}}|_{c^{l-1}_{12} = 1} 
\end{align}
A detailed derivation of the equivalence of  \eqref{eq:chi} and \eqref{eq:Cprime} is given in Appendix \ref{SM:jacobian_correlation}.
When there exists a fixed point $q^*$ such that $\mathcal{V}(q^*) = q^*$, and $\chi < 1$, then inputs with small initial correlation converge to correlation 1 at an exponential rate; this phase is referred to as \textit{ordered}. Alternatively, when $\chi > 1$ the fixed point $c^*=1$ becomes unstable, meaning that an input and its arbitrarily small perturbation have correlation $q^{l}_{ab}$ decreasing with layers; this is referred to as the \textit{chaotic} phase due to all nearby points on a data manifold diverging as they progress through the network. In the ordered phase, the output function of the network is constant whereas in the chaotic phase it is non-smooth everywhere. 

In both cases ($\chi>1$ or $\chi<1$), in \cite{Schoenholz_2017}, the mappings $\mathcal{V}$ and $\frac{1}{q^*}\mathcal{C}$ are shown to converge exponentially fast to their fixed point, when they exist. Therefore, the data geometry is quickly lost as it is propagated through layers.  The boundary between these phases, where $\chi=1$, is referred to as the edge-of-chaos and determines the scaling of $(\sigma_w,\sigma_b,\gamma_l)$, as functions of nonlinear activation $\phi(\cdot)$, which ensures a sub-exponential asymptotic behaviour of these maps towards their fixed point and thus a deeper data propagation along layers which facilitates early training of the network.

Second, the quantity $\chi_{\gamma}$ in \eqref{eq:chi} is equal to the median singular value of the the matrix $D^{(l)}W^{(l)}$ where $D^{(l)}$ is the diagonal matrix with at layer $l$ with entries $D_{ii}^{(l)}=\phi'(h_i^l)$; for details see Appendix \ref{SM:average_singular_value}.  Defining the Jacobian matrix $J \in \mathbb{R}^{N_L \times N_0}$ of the input-output map as 
\begin{align}\label{eq:J}
    J := \frac{\partial z^L}{\partial z^0} = \prod_{l=1}^L D^{(l)} W^{(l)},
\end{align}
we see that the average singular value of $J$ is equal to $\chi_{\gamma}^L$.  If $\chi_{\gamma}=1$ the average singular value of $J$ is fixed at 1 throughout the network, while if $\chi_{\gamma}$ is greater than or less than 1 the average singular value deviates from 1 at an exponential rate. Further note that the growth of a perturbation from a layer to the following one is given by the average squared singular value of $D^{(l)} W^{(l)}$.

\subsection{Main contributions}
This manuscript extends the edge-of-chaos analysis of random feed-forward networks to the setting of low-rank matrices, following the work of \cite{Poole_2016}. This work is motivated by the recent challenges faced to store in memory the constantly growing number of parameters used to train large Deep Learning models, see \cite{Price_2023} and references therein.  

\begin{figure}[ht]
    \vskip 0.2in
    \begin{center}
\centerline{\includegraphics[width=0.95\columnwidth]{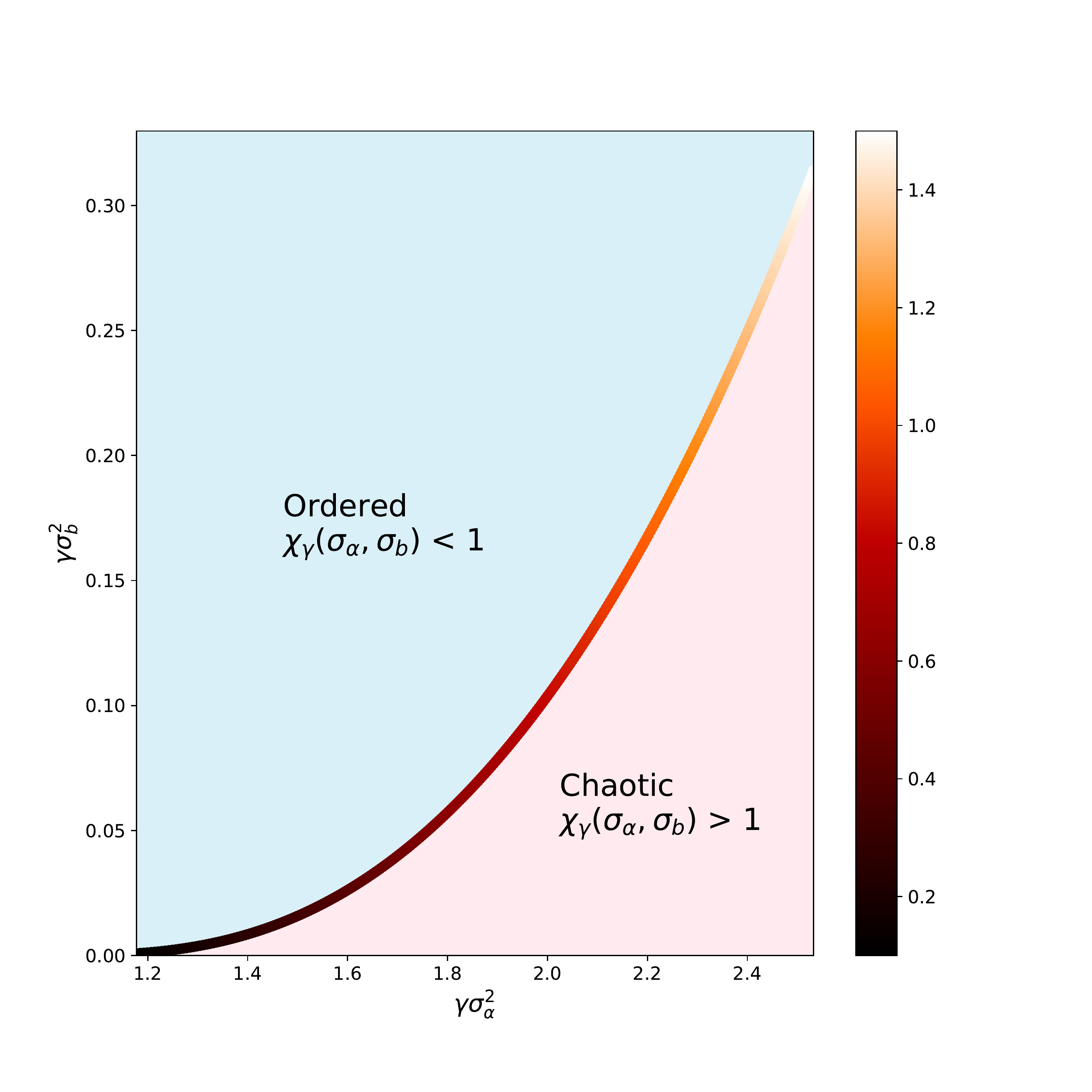}}
    \caption{Edge of Chaos curve of a low-rank neural network where the rank is proportional to the width by a factor $\gamma$ and nonlinear activation $\phi(x)=tanh(x)$ . The plot is generated with $\gamma=\frac{1}{4}$, where the axis are rescaled by $\gamma$.}
    \label{fig:EOC_gamma14}
    \end{center}
    \vskip -0.2in
\end{figure}

As shown in equations \eqref{eq:qmap}, \eqref{eq:Cmap}, and \eqref{eq:chi}, despite the dependence between entries in the low-rank weight matrices \eqref{definition_weights}, that the edge-of-chaos curve defined by $\chi_{\gamma}=1$ can be retained by scaling the weight and bias variances $\sigma_w^2$ and $\sigma_b^2$ respectively by the ratio of the weight matrix rank $r_l$ to layer width $\gamma_l:=r_l/N_l$, see Figure \ref{fig:EOC_gamma14} and contrast with Figure \ref{fig:EOC_gamma1}.  That is, a simple re-scaling retains the dominant first order dynamics of a feedforward network when the weight matrices are initialized to be low-rank.

In Section \ref{sec2:proportional_rank} we show that additional first order dynamics are similarly modified through a multiplicative scaling by the rank to width factor $\gamma_l=r_l/N_l$.  In particular, we demonstrate the role of $\gamma_l$ on the length and correlation depth scale as well as  the training gradient vectors.

However, in Section \ref{sec:dynamical_isometry} we show that important second order properties of the dynamics, specifically the variance of the singular values of the input-output Jacobian given in \eqref{eq:J}, is modified by the reduced rank in a way that cannot be overcome with simple re-scaling. This result alerts practitioners to anticipated greater variability in training low-rank weight matrices and suggests that methods to reduce the variance of the spectrum may be increasingly important in this setting, see \cite{Murray_2021}.

The manuscript then concludes with numerical experiments in Section \ref{sec:numerical_experiments} which demonstrate that empirical measurements on the Jacobian are consistent with the established formula and a brief summary and future work in Section \ref{sec:summary}.

\section{Network dynamics and data propagation}
\label{sec2:proportional_rank}

 The parameter $\chi_{\gamma}$ further controls the length and correlation depth scaling as well as the relative magnitude of training gradients computed via back-propagration for the sum-of-squares loss function.

\subsection{Depth scales a functions of $\chi_{\gamma}$}

The role of $\chi_{\gamma}$ on the achievable depth scale was pioneered by \cite{Schoenholz_2017} for full-rank feedforward networks.  In this subsection we extend their results to the low-rank setting with the suitably adapted spectral mean $\chi_{\gamma}$ given in \eqref{eq:chi}.  

\subsubsection{Length depth scale}
Assuming there exists a fixed point $q^*$ such that $\mathcal{V}(q^*) = q^*$, then the dynamics of $\mathcal{V}(q)$ can be linearized around $q^*$ to obtain stability conditions and a rate of decay which determine how deeply data can propagate through the network before converging towards the fixed point. 
Following the computations done in \cite{Schoenholz_2017}, setting a perturbation around the fixed point $q^* + \epsilon_l$, then around the fixed point, $\epsilon_l$ evolves as $e^{-\frac{l}{\xi_{q,\gamma}}}$, when $\gamma_l='gamma$ is fixed along layers and we define the following quantities,

\begin{align*}
\xi_{q,\gamma}^{-1} &:= - \log\bigg(\chi_{\gamma} + \gamma \sigma^2_{\alpha} \int Dz \phi^{\prime\prime}(\sqrt{q^*}z)\phi(\sqrt{q^*}z)\bigg).
\end{align*}
Details are given in \cref{SM:length_scale}. Given that $\gamma \in (0,1]$, we can see the convergence gets faster towards the fixed point when increasing $\gamma$. Note that when $\gamma \sigma^2_{\alpha} = \sigma^2_W$, we recover the results of a full-rank feedforward neural network in \cite{Schoenholz_2017}. %Because one may want to normalise the data beforehand as a preprocessing step, we will analyse the propagation of the input pairwise correlation as it seems to us to be a more relevant feature of the network's ability to propagate signals through depth.

\subsubsection{Correlation depth scale}

Similarly, we compute the dynamical evolution of the correlation map around its fixed point by considering a perturbation $\epsilon_l$ and we obtain that (see \cref{SM:correlation_scale}), when all the ranks are set to be proportional to the width with the same coefficient of proportionality $\gamma_l = \gamma$ at any layer $l$, then the perturbation vanishes exponentially fast $\epsilon_l = \mathcal{O}(e^{-\frac{l}{\xi_{q, \gamma}}})$ where 
\begin{align*}
\xi_{c,\gamma}^{-1} &:= - \log \big(\chi_{\gamma} \big).
\end{align*}

We recover that the correlation depth scale diverges to $+\infty$ when $\chi_{\gamma} \to 1$, yielding again the key role of this quantity, even in the low-rank case. As $\gamma \in (0,1]$, we can see the convergence gets faster towards the fixed point when increasing $\gamma$, which highlights the tension between low-rank and the depth to which data can propagate along layers. Note again we recover previous results from \cite{Schoenholz_2017} after appropriate scaling of the variance. 

\subsection{Layerwise scaling of the training gradient for the sum-of-squares loss function}
As already shown in previous works (\cite{Schoenholz_2017}, and \cite{Poole_2016}), there exists an direct link between the capacity for a network to propagate data through layers of a network in the forward pass and to backpropagate gradients of any given error function $E$. In this section, we extend the results known for full-rank feedforward neural networks with infinite width to the low-rank case, with rank $r_l=\gamma_l N_l$ evolving proportionally to the width.

The derivative of the training error follows by the chain rule,
\begin{align*}
&\frac{\partial E}{\partial h_i^{(l)}} := \delta_i^l = \big(\sum\limits_{k=1}^{N_{l+1}} \delta_k^{l+1} W_{ki}^{(l+1)} \big) \phi^\prime(h_i^{(l)}),\\
&\frac{\partial E}{\partial W_{ij}^{(l)}} = \delta_i^l \phi(h_j^{(l-1)}),\\
&\frac{\partial E}{\partial \alpha_{ij}^{(l)}} = \bigg(\sum\limits_{m=1}^{r_l} \delta_m^l (C_i^l)_m\bigg) \phi(h_j^{(l-1)}).
\end{align*}

Consider the propagation of the gradients $\frac{\partial E}{\partial \alpha_{ij}^{(l)}}$ of the error with respect to our trainable parameters $\alpha^{(l)} := \big(\alpha^{(l)}_{i,j}\big)_{i,j}$, which are initalized $\left(\alpha^{(l)}_{k,i}\right)_{1\leq i \leq N_{l-1}}\in \mathbb{R} \overset{\text{iid}}{\sim} \mathcal{N}(0,\frac{\sigma_{\alpha}^2}{N_{l-1}})$. The length of this gradient along layers $||\nabla_{\alpha^{(l)}} E||^2_2$ is proportional to ${\tilde{q}}^l:=\mathbb{E}((\delta_1^l)^2)$ (see \cref{SM:backpropagation} for proofs).  In our analysis of the variance of the training error we treat the backpropagated weights as independent from the forwarded weights, which wile not strictly true is commonly done due to its efficacy in aiding computations which reflect the observed backward dynamics of the network, see \cite{Pennington_2017_geometry}. Considering an input vector $x^{0,a}$, and ${\tilde{q}}^l_{aa}:={\tilde{q}}^l(x^{0,a})$, 
\begin{align*}
{\tilde{q}}^l_{aa} = {\tilde{q}}^{l+1}_{aa} \frac{N_{l+1}}{N_l} \chi_{l+1},
\end{align*}
see \cref{SM:backpropagation}.

With constant width along layers $\frac{N_{l+1}}{N_l} \approx 1$, then the sequence is asymptotically exponential and ${\tilde{q}}^l_{aa} = {\tilde{q}}^L_{aa} \prod_{k=l+1}^{L} \chi_{\gamma_k}$, or, if the proportional coefficient of the rank $\gamma_l = \gamma$ is constant along layers, ${\tilde{q}}^l_{aa} = \mathcal{O}(e^{\frac{l}{\xi_{\Delta,\gamma}}})$, where
\begin{align*}
\xi_{\Delta,\gamma}^{-1} &:= - \log(\chi_{\gamma}) 
\end{align*}
The same critical point is observed in the low-rank setting $\gamma<1$ as in previous works \cite{Schoenholz_2017} given by $\chi_{\gamma} = 1$ :
\begin{itemize}
    \item When $\chi_{\gamma}> 1$, then $||\nabla_{\alpha^{(l)}} E||^2_2$ grows exponentially after $|\xi_{\nabla, \gamma}|$ layers. This is the chaotic phase with the network is being exponentially-sensitive to perturbations.
    \item When $\chi_{\gamma} < 1$, then $||\nabla_{\alpha^{(l)}} E||^2_2$ vanishes at an exponential rate after $\xi_{\nabla, \gamma}$ layers. This is the ordered phase with the network is being insensitive to perturbations.
    \item When $\chi_{\gamma} = 1$, then $||\nabla_{\alpha^{(l)}} E||^2_2$ remains of the same scale across even after an infinite number of layers which is referred to as the edge-of-chaos.
\end{itemize}

\section{Dynamical isometry}\label{sec:dynamical_isometry}
%Similarly to some prior works on controlling the variance of the Jacobian'spectrum and avoiding any exponential growth or shrinkage with depth of gradients, especially on the Edge of Chaos (e.g. the mean of the spectra is set to be 1), we extend the known results to our low-rank setting. This property of having all the Jacobian'spectra concentrated at 1 is referred as Dynamical isometry. More specifically we will investigate the impact of the rank on the shown benefit from using orthogonal weights rather than Gaussian when initialising the network. 
Using tools from Random Matrix Theory, \cite{Pennington_2018_emergence} provides a method to compute the moments of the spectral distribution of the Jacobian, revealing secondary information beyond the mean of the spectra.
We review the most essential equations to derive the variance of the Jacobian'spectrum here but we refer the reader to \cite{Random_matrix_book} for more details on the random matrix transforms.

\subsection{Review of the computation of the variance of the Jacobian}

In this section, we review a set of definitions of random matrix transforms that allow the calculation of the spectra of the product of matrices in terms of their individual spectra.  Let $X$ be a random matrix with spectral density $\rho_X$
\begin{align*}
\rho_X(\lambda) := \langle \frac{1}{N}\sum\limits_{i=1}^N \delta(\lambda - \lambda_i)\rangle_{X},
\end{align*}
where $\langle . \rangle$ is the average with respect to the distribution of the random matrix $X$, and $\delta$ is the usual dirac distribution.

For a probability density $\rho_X$ and $z\in \mathbb{C}\setminus \mathbb{R}$, the Stieltjes Transform $G_X$ and its inverse are given by
\begin{align*}
G_X(z) :&= \int \frac{\rho_X(t)}{t-z} dt,\\
\rho_X(\lambda) &= -\pi^{-1} \underset{\epsilon \to 0^+}{\lim} \text{Im}\big(G_X(\lambda+\epsilon i)\big).
\end{align*}
The moment generating function is $M_X(z):=zG_x(z)-1 = \sum\limits_{k=1}^\infty m_k z^{-k}$ and the $\mathcal{S}_X$ Transform is defined as $\mathcal{S}_X(z) := \frac{1+z}{xM_X^{-1}(z)}$. The interest of using the $\mathcal{S}$ Transform here is that it has the following multiplicative property, which in our case is desirable as the Jacobian is a product of random matrices: if $X$ and $Y$ are freely independent, then $\mathcal{S}_{XY} = \mathcal{S}_X \mathcal{S}_Y$. 

In \cite{Pennington_2018_emergence}, the authors start with establishing $\mathcal{S}_{JJ^T} = S_{D^2}^L S_{W^TW}^L$, assuming the input vector is chosen such that $q^l\approx q^*$ so that distribution of $D^2$ is independent of $l$ and we already had the weights identically distributed along layers. The strategy here to compute the spectral density of $\rho_{JJ^T}$ (and thus the density of the singular values of the Jacobian $J$) starts with computing the $\mathcal{S}$ Transforms of $W^TW$ and $D^2$ from their spectral density, determined by respectively, the way of sampling the weights at initialisation and the choice of the activation function in the network. Note that in this study we focus only on two possible distributions for the low-rank weights matrix - either scaled Gaussian weights or scaled orthogonal matrices, that are defined more precisely in the next sections. Once that $\mathcal{S}_{JJ^T}$ is obtained by multiplying $S_{W^TW}$ and $S_{D^2}$, rather than inverting it back to find $\rho_{JJ^T}$, the authors show there is a way to shortcut these steps and obtain directly the moments of $\rho_{JJ^T}$ based on the following set of equations. Defining
\begin{align}
m_k &:= \int \lambda^k \rho_{JJ^T}(\lambda)d\lambda \nonumber\\
S_{W^TW}(z) &:= \gamma^{-1} \sigma^{-2}_{\alpha} \big(1+\sum\limits_{k=1}^\infty s_k z^k\big)\nonumber\\
\mu_k &= \int Dz \big(\phi^\prime(\sqrt{q^*}z)\big)^{2k} \nonumber\\
\intertext{then as derived in \cite{Pennington_2018_emergence}, the first two moments of the spectrum of the Jacobian are}
m_1 &= (\gamma \sigma_{\alpha}^2 \mu_1)^L \nonumber\\
m_2 &= (\gamma \sigma_{\alpha}^2 \mu_1)^{2L} L \left(\frac{\mu_2}{\mu_1^2} + \frac{1}{L} - 1 -s_1\right). \nonumber\\
\intertext{The first moment $m_1$ recovers the previous statement that the average squared singular value is equal to $m_1 = \chi_{\gamma}^L$ and the edge-of-chaos given by $\chi_{\gamma}=\gamma \sigma_{\alpha}^2 \mu_1=1$ is consistent with previous results as the gradient either vanishes or grows exponentially along with the median of the Jacobian's spectra. Moreover, the variance of the spectrum of $JJ^T$ about its mean $\chi_{\gamma}=1$ can now be computed}
\sigma^2_{JJ^T} &:= m_2 - m_1^2 = L\left(\frac{\mu_2}{\mu_1^2} - 1 - s_1\right). \label{eq:variance_spectrum}
\end{align}
The variance $\sigma^2_{JJ^T}$ grows linearly with depth as in the  full-rank setting, recovering the full-rank result when $\gamma=1$. As in the edge-of-chaos axes scaling in Figure \ref{fig:EOC_gamma14}, $\gamma \sigma^2_{\alpha}$ plays the same role as $\sigma^2_W$. Note that $\frac{\mu_2}{\mu_1^2}\ge 1$ and consequently $\sigma^2_{JJ^T}$ as given in \eqref{eq:variance_spectrum} is only independent of depth $L$ if $s_1=0$ which is only achieved here in the case of full-rank, i.e. $\gamma=1$ orthogonal matrices.  

\subsection{Low-Rank Orthogonal weights}

Consider a weight matrix whose $r$ first columns are orthonormal columns sampled from a normal distribution, and the rest is 0, such that ${W}^TW  = \begin{pmatrix}
\sigma_{\alpha}^2 \mathbb{I}_{r} & \rvline & 0 \\
\hline
 0 & \rvline& \mathbb{O}_{N-r}
\end{pmatrix}$.
Therefore the spectral distribution of $\sigma_{\alpha}^{-2}{W}^TW$ is trivially given by
\begin{align*}
\rho_{\sigma_{\alpha}^{-2}{W}^TW}(z) = \gamma \delta(z-1) + (1-\gamma) \delta(z),
\end{align*}
from which the $\mathcal{S}$ Transform is computed, see \cref{SM:orthogonal_weights}, to obtain $s_1 = -(\gamma^{-1} - 1)$. When $\gamma=1$, the known result in the full-rank orthogonal case is retrieved. 
%See \cref{SM:orthogonal_weights} for a detailed computed of the $\mathcal{S}$ Transform in this case.

\subsection{Low-Rank Gaussian weights}
With weights at any layer $l$ given by \ref{definition_weights}, the matrix can be rewritten as the product $W^l = C^l A^l $, where $C^l \in \mathbb{R}^{N_l \times r_l}$ with $C^l_{ij} = (C_j^l)_i$, and $A^l \in \mathbb{R}^{r_l \times N_{l-1}}$ with $A^l_{ij} \overset{iid}{\sim}\mathcal{N}(0,\frac{\sigma^2_{\alpha}}{N_{l-1}})$. As ${C^l}^TC^l = \mathbb{I}_{r_l}$ by construction, then ${W^{(l)}}^T W^{(l)} = {A^l}^T {C^l}^T C^l A^l = {A^l}^T A^l$ which is a Wishart matrix, whose spectral density is known and given by the Marčenko Pastur distribution \cite{Marchenko_1967} where some mass is added at 0 since the matrix $A^l$ is not full-rank and contains some 0 eigenvalues. Recall that $r_l = \gamma N_l$.

\begin{align*}
\medmath{\rho_{A^TA}(\lambda)} = \medmath{(1 - \gamma)_+ \delta(\lambda) + \gamma \frac{\sqrt{(\lambda^+ - \lambda)(\lambda - \lambda^-)}}{2\pi\lambda \sigma_{\alpha}^2} \mathds{1}_{[\lambda^-,\lambda^+]}(\lambda)},
\end{align*}
where $x_+ = \text{max}(0,x), \lambda^- = (1-\frac{1}{\gamma})^2$ and $\lambda^+ = (1+\frac{1}{\gamma})^2$.
The $\mathcal{S}$ Transform $S_{W^TW}$ can be computed (see \cref{SM:Gaussian_weights}) and expanded around 0, which gives $s_1 = -\frac{1}{\gamma}$. Note that when $\gamma=1$, one recovers the result given in \cite{Pennington_2018_emergence}.

The $\mathcal{S}$ Transforms and first moments in both orthogonal and Gaussian cases are summarized in \cref{table:transform-weights}.

% table with recap of the value of s_1 in orthogonal/Gaussian weights case

\begin{table}[t]
\caption{Transforms of weights. LR stands for Low-Rank.}
\label{table:transform-weights}
\vskip 0.15in
\begin{center}
\begin{tiny}
\begin{sc}
\begin{tabular}{lcccr}
\toprule
Random Matrix W & $S_{W^TW}(z)$ & $s_1$ \\
\midrule
LR Scaled Orthogonal    & $\gamma^{-1} \sigma_{\alpha}^{-2} \frac{1+z} {1+\gamma^{-1}z}$ & $1-\frac{1}{\gamma}$ \\
LR Scaled Gaussian & $\gamma^{-1}\sigma_{\alpha}^{-2} \frac{1+z}{1 + z(1+\gamma^{-1}) + \gamma^{-1} z^2}$ & $-\frac{1}{\gamma}$ \\
\bottomrule
\end{tabular}
\end{sc}
\end{tiny}
\end{center}
\vskip -0.1in
\end{table}

\section{Numerical experiments}
\label{sec:numerical_experiments}

\begin{figure}[ht]
    \vskip 0.2in
    \begin{center}
\centerline{\includegraphics[width=\columnwidth]{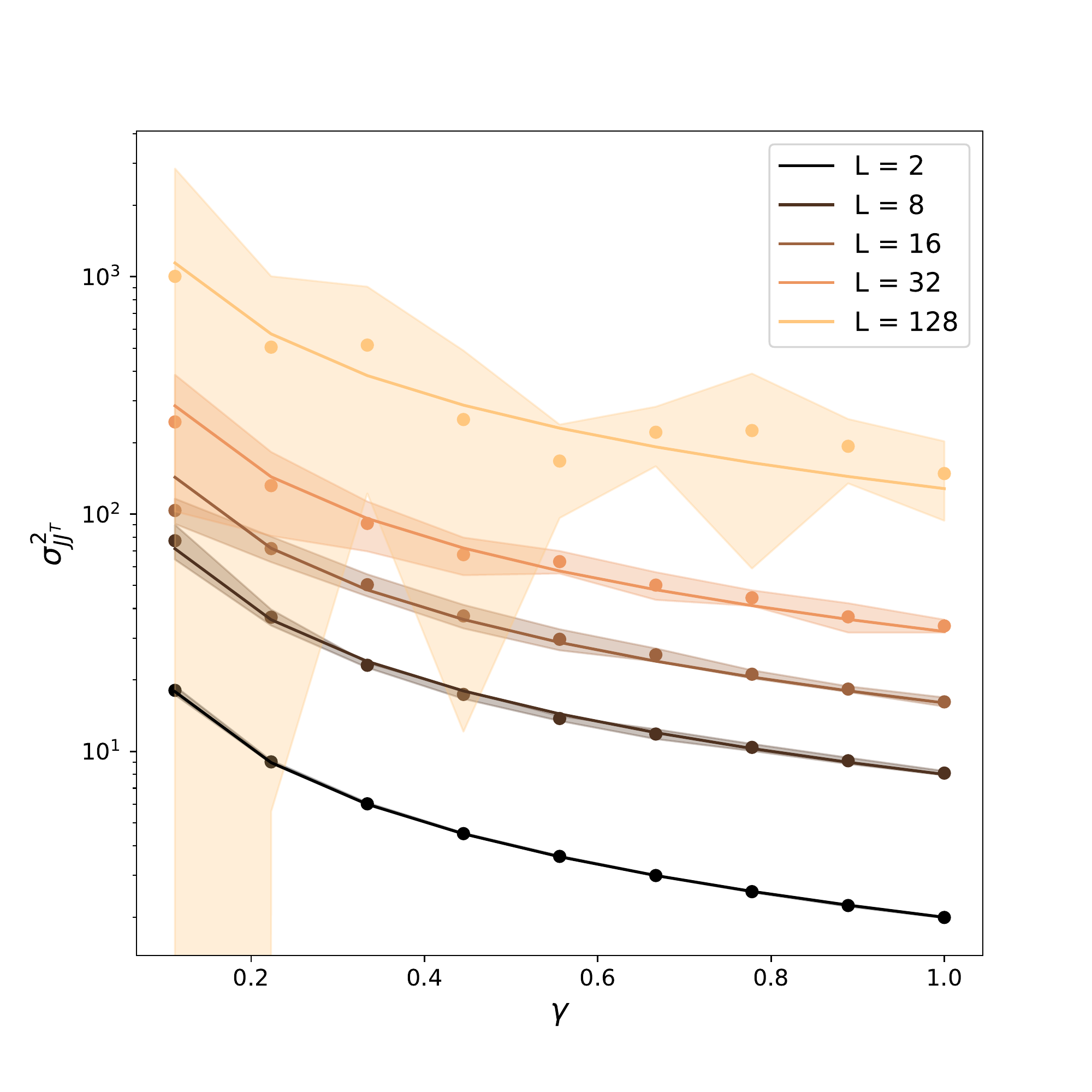}}
    \caption{Evolution of the variance of the spectrum of $JJ^T$ with respect to $\gamma$ where $\gamma$ is the proportionality coefficient giving the rank of the weights matrices at layer $l$, whose width is $N_l$, $r_l = \gamma N_l$. Points are obtained empirically and averaged over 5 simulations when the lines are derived from the theory, see \eqref{eq:variance_spectrum}. Confidence intervals of 1 standard deviation around each mean point are shown. The weights are chosen to be low-rank Scaled Gaussian and the activation function is linear $\phi:x\mapsto x$. The same seed is used to initialise the weight matrices for each simulation and $q^*$ is set to 0.5. The $y-$axis is shown in log scale.}
    \label{fig:variance_spectrum_linear_Gaussian}
    \end{center}
    \vskip -0.2in
\end{figure}

\begin{figure}[ht]
    \vskip 0.2in
    \begin{center}
\centerline{\includegraphics[width=\columnwidth]{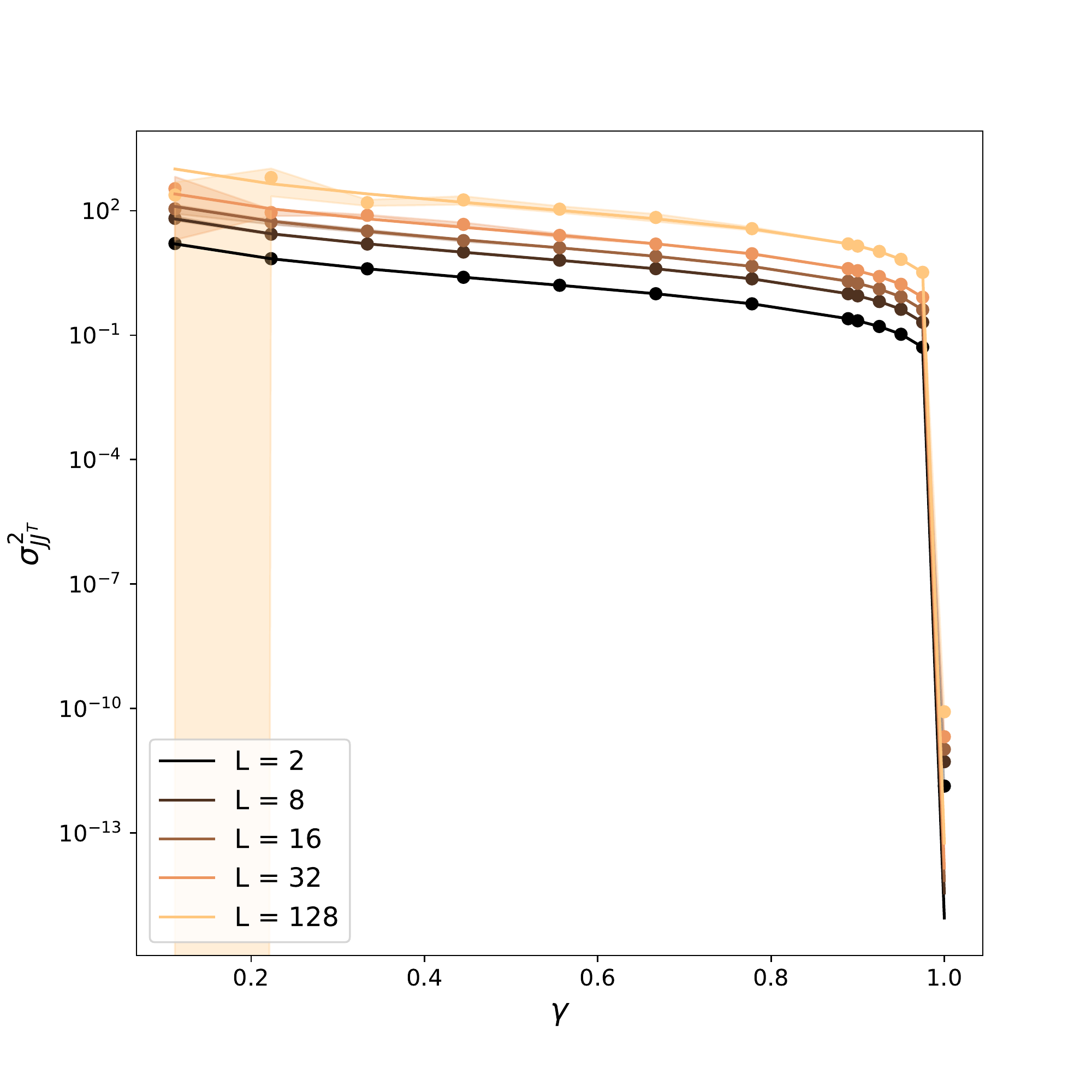}}
    \caption{Evolution of the variance of the spectrum of $JJ^T$ with respect to $\gamma$ where $\gamma$ is the proportionality coefficient giving the rank of the weights matrices at layer $l$, whose width is $N_l$, $r_l = \gamma N_l$. Points are obtained empirically and averaged over 3 simulations when the lines are derived from the theory, see \eqref{eq:variance_spectrum}. Confidence intervals of 1 standard deviation around each mean point are shown. The weights are chosen to be low-rank Scaled Orthogonal and the activation function is linear $\phi:x\mapsto x$. The same seed is used to initialise the weight matrices for each simulation and $q^*$ is set to 0.5. The $y-$axis is shown in log scale.}
    \label{fig:variance_spectrum_orthogonal_Gaussian_log_scale}
    \end{center}
    \vskip -0.2in
\end{figure}

In this section, we give empirical evidence in agreement with the theoretical results established above. Its interest is two-fold:
\begin{itemize}
    \item The variance of the spectrum of the Jacobian does indeed still grow with depth even in the low-rank setting as emphasized in \cref{fig:variance_spectrum_linear_Gaussian}. Moreover, at a fixed depth, the rank to width ratio plays a key role in how the spectrum of the Jacobian spreads out around its mean value, which is 1 when the network is initialised on the edge-of-chaos.
    \item  \cref{fig:variance_spectrum_orthogonal_Gaussian_log_scale} shows that the advantage that Scaled Orthogonal Weights have over Scaled Gaussian Weights in Feedforward networks presented in \cite{Pennington_2018_emergence} is lost for low-rank matrices. Indeed, from \eqref{eq:variance_spectrum}, one can see that in both situations, it is not possible  to adjust either the activation function nor $q^*$ through a careful choice of variances for the weights and the biases, unless $\gamma=1$ and $W^{(l)}$ is a scaled orthonormal matrix.
\end{itemize}

In \cref{fig:variance_spectrum_linear_Gaussian} and \cref{fig:variance_spectrum_orthogonal_Gaussian_log_scale}, the variance of the spectrum of the Jacobian is computed in the low-rank Gaussian and Orthogonal cases when the activation function is chosen to be the identity. Although such a choice of activation function completely destroys the network's expressivity power, it is a simple example of situations in the full-rank case where Gaussian distributed weight matrices lead to ill conditioned Jacobians as depth increases. This still holds in the low-rank setting as shown in the plot since the variance $\sigma^2_{JJ^T}>0$. Simulations are performed on a $1000$- layer wide feedforward network, initialised and fed with a random input, whose length is set to be equal to $q^*$ so that the network would already be at its equilibrium state without passing by a transient phase.

The source code can be found at \url{shorturl.at/syLP9}.

\begin{figure}[ht]
    \vskip 0.2in
    \begin{center}
\centerline{\includegraphics[width=1.2\columnwidth]{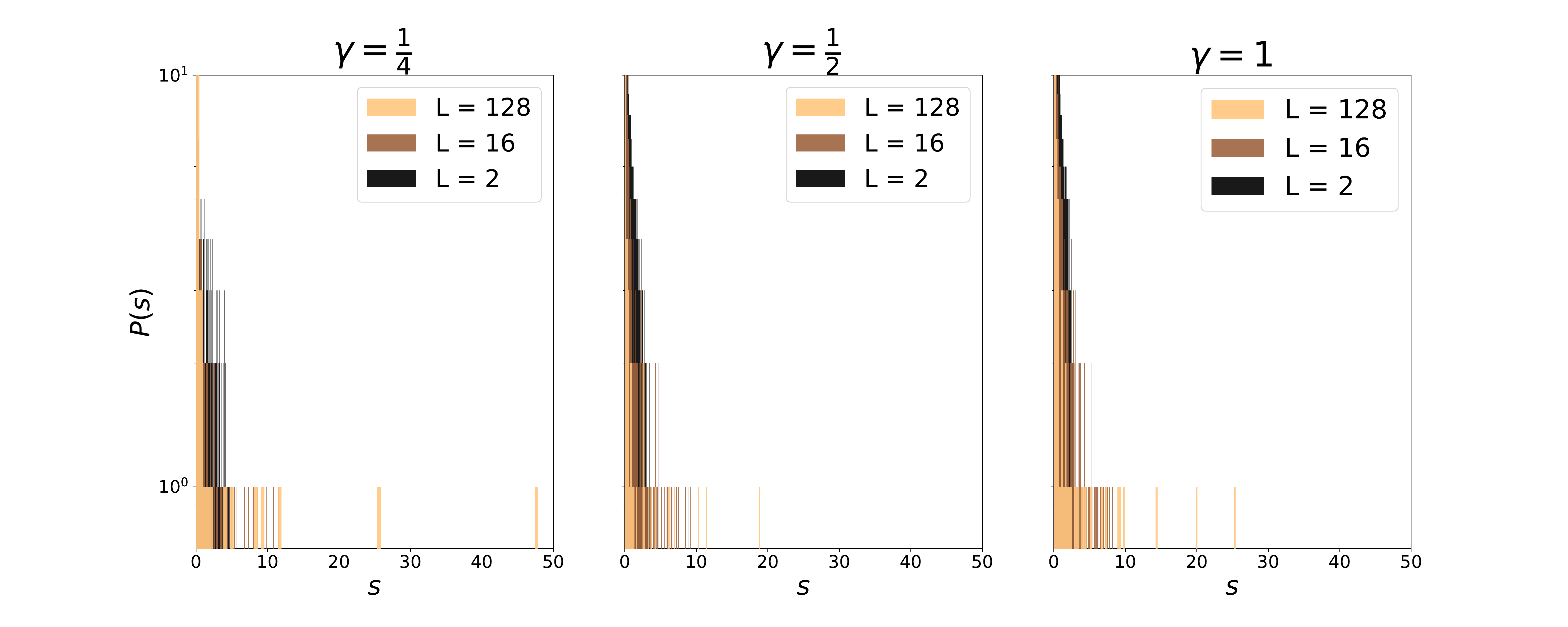}}
    \caption{Singular values of the Jacobian $J$ with respect to the depth of the network, whose weight matrices are low-rank Scaled Gaussian. The rank to width ratio $\gamma$ increases on each plot from left to right when the width is kept constant to $1000$. The activation function is $\text{erf}$. The same seed is used to initialise the weight matrices for each simulation and $q^*$ is set to 0.5. The $y-$axis is shown in log scale.}
    \label{fig:entire_emp_spectrum_Gaussian_erf}
    \end{center}
    \vskip -0.2in
\end{figure}

\begin{figure}[ht]
    \vskip 0.2in
    \begin{center}
\centerline{\includegraphics[width=1.2\columnwidth]{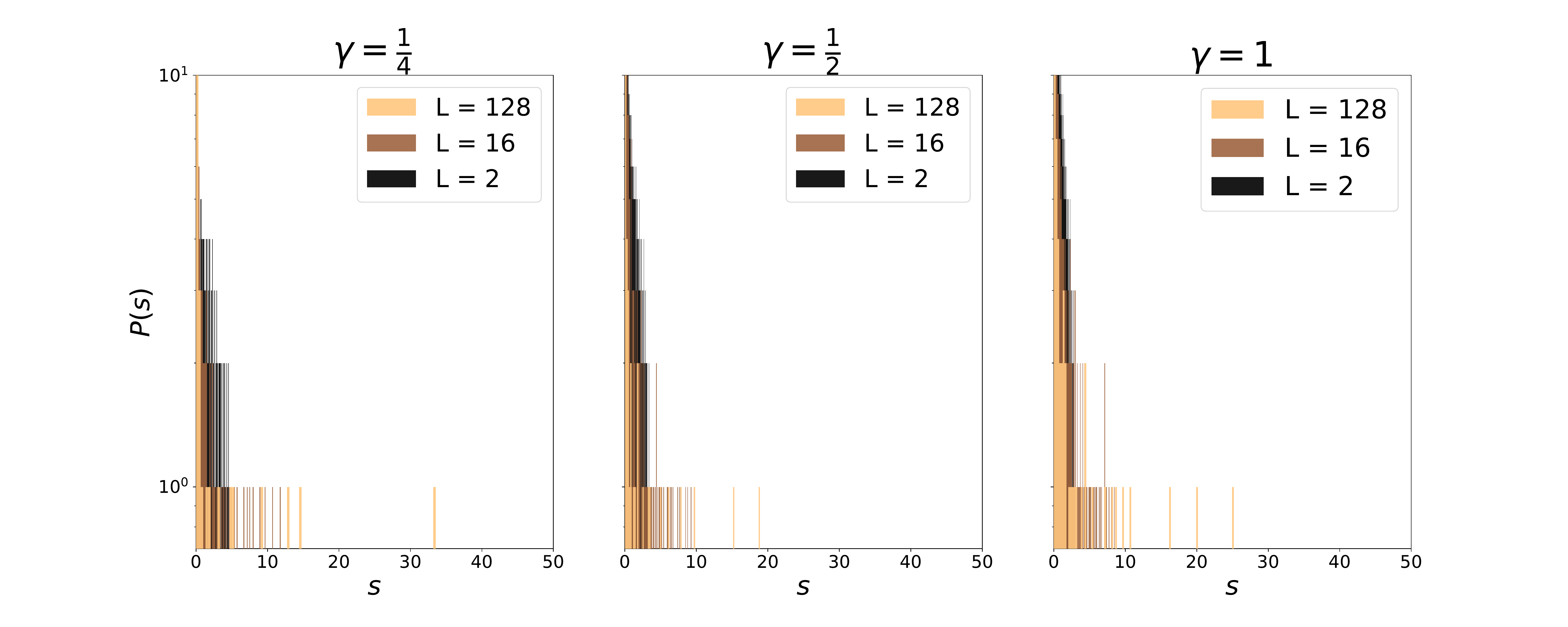}}
    \caption{Singular values of the Jacobian $J$ with respect to the depth of the network, whose weight matrices are low-rank Scaled Gaussian. The rank to width ratio $\gamma$ increases on each plot from left to right when the width is kept constant to $1000$. The activation function is $\text{tanh}$. The same seed is used to initialise the weight matrices for each simulation and $q^*$ is set to 0.5. The $y-$axis is shown in log scale.}
    \label{fig:entire_emp_spectrum_Gaussian_tanh}
    \end{center}
    \vskip -0.2in
\end{figure}

\section{Summary and further work}
\label{sec:summary}

Herein the edge-of-chaos theory of \cite{Poole_2016} and \cite{Schoenholz_2017} has been extended from the setting of full-rank weight matrices to the low-rank setting.  Suitable scaling by the ratio of the rank to width factor $\gamma_l:=r_l/N_l$ recovers the phenomenon driven by the mean of the Jacobian's spectra which defines the edge-of-chaos.  Moreover, the variance of the Jacobian's spectra is shown to be strictly increasing with decreasing $\gamma_l$ which suggests greater variability in the initial training of low-rank feedforward networks.

% Second paragraph
The edge-of-chaos initialisation scheme has been successfully generalised to a large set of different settings, including changes of architectures as CNNs \cite{Xiao_2018}, LSTMs and GRUs \cite{Gilboa_2019}, RNNs \cite{Chen_2018}, ResNets \cite{Yang_2017} and to extra features like dropout \cite{Schoenholz_2017}, \cite{Huang_2019} or batch normalisation \cite{Yang_2019} and pruning \cite{Hayou_2020}. It has been improved with changes of activation functions \cite{Hayou_2019}, \cite{Murray_2021} to enable the data to propagate even deeper through the network.  As a future work, each of these settings could be extended to the setting of low-rank weight matrices.

\section*{Acknowledgments} 
TNS is financially supported by the Engineering and Physical Sciences Research Council (EPSRC). JT is supported by the Hong Kong Innovation and Technology Commission (InnoHK Project CIMDA) and thanks UCLA Department of Mathematics for kindly hosting him during the completion of this manuscript.

% In the unusual situation where you want a paper to appear in the
% references without citing it in the main text, use \nocite
%\nocite{langley00}

\bibliography{example_paper}

\begin{thebibliography}{17}
\providecommand{\natexlab}[1]{#1}
\providecommand{\url}[1]{\texttt{#1}}
\expandafter\ifx\csname urlstyle\endcsname\relax
  \providecommand{\doi}[1]{doi: #1}\else
  \providecommand{\doi}{doi: \begingroup \urlstyle{rm}\Url}\fi

\bibitem[Chen et~al.(2018)Chen, Pennington, and Schoenholz]{Chen_2018}
Chen, M., Pennington, J., and Schoenholz, S.
\newblock Dynamical isometry and a mean field theory of {RNN}s: Gating enables
  signal propagation in recurrent neural networks.
\newblock In Dy, J. and Krause, A. (eds.), \emph{Proceedings of the 35th
  International Conference on Machine Learning}, volume~80 of \emph{Proceedings
  of Machine Learning Research}, pp.\  873--882. PMLR, 10--15 Jul 2018.
\newblock URL \url{https://proceedings.mlr.press/v80/chen18i.html}.

\bibitem[Gilboa et~al.(2019)Gilboa, Chang, Chen, Yang, Schoenholz, Chi, and
  Pennington]{Gilboa_2019}
Gilboa, D., Chang, B., Chen, M., Yang, G., Schoenholz, S.~S., Chi, E.~H., and
  Pennington, J.
\newblock Dynamical isometry and a mean field theory of lstms and grus, 2019.
\newblock URL \url{https://arxiv.org/abs/1901.08987}.

\bibitem[Glorot \& Bengio(2010)Glorot and Bengio]{Glorot_2010}
Glorot, X. and Bengio, Y.
\newblock Understanding the difficulty of training deep feedforward neural
  networks.
\newblock In Teh, Y.~W. and Titterington, M. (eds.), \emph{Proceedings of the
  Thirteenth International Conference on Artificial Intelligence and
  Statistics}, volume~9 of \emph{Proceedings of Machine Learning Research},
  pp.\  249--256, Chia Laguna Resort, Sardinia, Italy, 13--15 May 2010. PMLR.
\newblock URL \url{https://proceedings.mlr.press/v9/glorot10a.html}.

\bibitem[Hayou et~al.(2019)Hayou, Doucet, and Rousseau]{Hayou_2019}
Hayou, S., Doucet, A., and Rousseau, J.
\newblock On the impact of the activation function on deep neural networks
  training, 2019.
\newblock URL \url{https://arxiv.org/abs/1902.06853}.

\bibitem[Hayou et~al.(2020)Hayou, Ton, Doucet, and Teh]{Hayou_2020}
Hayou, S., Ton, J.-F., Doucet, A., and Teh, Y.~W.
\newblock Robust pruning at initialization, 2020.
\newblock URL \url{https://arxiv.org/abs/2002.08797}.

\bibitem[Huang et~al.(2019)Huang, Da~Xu, Du, Zeng, and Zhao]{Huang_2019}
Huang, W., Da~Xu, R.~Y., Du, W., Zeng, Y., and Zhao, Y.
\newblock Mean field theory for deep dropout networks: digging up gradient
  backpropagation deeply.
\newblock 2019.
\newblock \doi{10.48550/ARXIV.1912.09132}.
\newblock URL \url{https://arxiv.org/abs/1912.09132}.

\bibitem[Marčenko \& Pastur(1967)Marčenko and Pastur]{Marchenko_1967}
Marčenko, V.~A. and Pastur, L.~A.
\newblock Distribution of eigenvalues for some sets of random matrices.
\newblock \emph{Mathematics of the USSR-Sbornik}, 1\penalty0 (4):\penalty0 457,
  apr 1967.
\newblock \doi{10.1070/SM1967v001n04ABEH001994}.
\newblock URL \url{https://dx.doi.org/10.1070/SM1967v001n04ABEH001994}.

\bibitem[Murray et~al.(2021)Murray, Abrol, and Tanner]{Murray_2021}
Murray, M., Abrol, V., and Tanner, J.
\newblock Activation function design for deep networks: linearity and effective
  initialisation, 2021.
\newblock URL \url{https://arxiv.org/abs/2105.07741}.

\bibitem[Pennington \& Bahri(2017)Pennington and
  Bahri]{Pennington_2017_geometry}
Pennington, J. and Bahri, Y.
\newblock Geometry of neural network loss surfaces via random matrix theory.
\newblock In Precup, D. and Teh, Y.~W. (eds.), \emph{Proceedings of the 34th
  International Conference on Machine Learning}, volume~70 of \emph{Proceedings
  of Machine Learning Research}, pp.\  2798--2806. PMLR, 06--11 Aug 2017.
\newblock URL \url{https://proceedings.mlr.press/v70/pennington17a.html}.

\bibitem[Pennington et~al.(2018)Pennington, Schoenholz, and
  Ganguli]{Pennington_2018_emergence}
Pennington, J., Schoenholz, S.~S., and Ganguli, S.
\newblock The emergence of spectral universality in deep networks, 2018.
\newblock URL \url{https://arxiv.org/abs/1802.09979}.

\bibitem[Poole et~al.(2016)Poole, Lahiri, Raghu, Sohl-Dickstein, and
  Ganguli]{Poole_2016}
Poole, B., Lahiri, S., Raghu, M., Sohl-Dickstein, J., and Ganguli, S.
\newblock Exponential expressivity in deep neural networks through transient
  chaos, 2016.
\newblock URL \url{https://arxiv.org/abs/1606.05340}.

\bibitem[Price \& Tanner(2022)Price and Tanner]{Price_2023}
Price, I. and Tanner, J.
\newblock Improved projection learning for lower dimensional feature maps,
  2022.
\newblock URL \url{https://arxiv.org/abs/2210.15170}.

\bibitem[Schoenholz et~al.(2016)Schoenholz, Gilmer, Ganguli, and
  Sohl-Dickstein]{Schoenholz_2017}
Schoenholz, S.~S., Gilmer, J., Ganguli, S., and Sohl-Dickstein, J.
\newblock Deep information propagation, 2016.
\newblock URL \url{https://arxiv.org/abs/1611.01232}.

\bibitem[Tao(2012)]{Random_matrix_book}
Tao, T.
\newblock \emph{Topics in Random Matrix Theory}.
\newblock Graduate studies in mathematics. American Mathematical Soc., 2012.
\newblock ISBN 9780821885079.
\newblock URL \url{https://books.google.co.uk/books?id=Hjq\_JHLNPT0C}.

\bibitem[Xiao et~al.(2018)Xiao, Bahri, Sohl-Dickstein, Schoenholz, and
  Pennington]{Xiao_2018}
Xiao, L., Bahri, Y., Sohl-Dickstein, J., Schoenholz, S., and Pennington, J.
\newblock Dynamical isometry and a mean field theory of {CNN}s: How to train
  10,000-layer vanilla convolutional neural networks.
\newblock In Dy, J. and Krause, A. (eds.), \emph{Proceedings of the 35th
  International Conference on Machine Learning}, volume~80 of \emph{Proceedings
  of Machine Learning Research}, pp.\  5393--5402. PMLR, 10--15 Jul 2018.
\newblock URL \url{https://proceedings.mlr.press/v80/xiao18a.html}.

\bibitem[Yang \& Schoenholz(2017)Yang and Schoenholz]{Yang_2017}
Yang, G. and Schoenholz, S.~S.
\newblock Mean field residual networks: On the edge of chaos, 2017.
\newblock URL \url{https://arxiv.org/abs/1712.08969}.

\bibitem[Yang et~al.(2019)Yang, Pennington, Rao, Sohl-Dickstein, and
  Schoenholz]{Yang_2019}
Yang, G., Pennington, J., Rao, V., Sohl-Dickstein, J., and Schoenholz, S.~S.
\newblock A mean field theory of batch normalization, 2019.
\newblock URL \url{https://arxiv.org/abs/1902.08129}.

\end{thebibliography}
\bibliographystyle{icml2022}

%%%%%%%%%%%%%%%%%%%%%%%%%%%%%%%%%%%%%%%%%%%%%%%%%%%%%%%%%%%%%%%%%%%%%%%%%%%%%%%
%%%%%%%%%%%%%%%%%%%%%%%%%%%%%%%%%%%%%%%%%%%%%%%%%%%%%%%%%%%%%%%%%%%%%%%%%%%%%%%
% APPENDIX
%%%%%%%%%%%%%%%%%%%%%%%%%%%%%%%%%%%%%%%%%%%%%%%%%%%%%%%%%%%%%%%%%%%%%%%%%%%%%%%
%%%%%%%%%%%%%%%%%%%%%%%%%%%%%%%%%%%%%%%%%%%%%%%%%%%%%%%%%%%%%%%%%%%%%%%%%%%%%%%
\newpage
\appendix
\onecolumn
\section{Supplementary Material}

\subsection{Preliminary lemma}
The following lemma is used later in the proofs contained in \cref{app:h_distribution}.
\begin{lemma}\label{lemma:convergence_norm_columns} Let $\gamma \in \mathbb{R}^*$.
If $C_1, \dots, C_{\gamma n} \overset{\text{iid}}{\sim} \mathcal{N}(0, \frac{1}{n})$, then $C_1^2 + \dots + C_{\gamma n}^2 \to \gamma $ in probability when $n\to \infty$.
\end{lemma}

\textit{Proof. }
If $X$ is a random variable, let us denote by $F_X$ its cumulative distribution function. Let $x \in \mathbb{R}$.
\begin{align*}
F_{C_1^2 + \dots + C_{\gamma n}^2}(x) & = \mathbb{P}(C_1^2 + \dots + C_{\gamma n}^2 \leq x)\\
& = \mathbb{P}(\sqrt{n} \frac{C_1^2 + \dots + C_{\gamma n}^2 - \gamma}{\sqrt{2\gamma}} \leq \sqrt{n} \frac{x-\gamma}{\sqrt{2\gamma}})\\
& =  \mathbb{P} (\frac{C_1^2 + \dots + C_{\gamma n}^2 - (n\gamma) \frac{1}{n}}{\frac{\sqrt{2}}{n} \sqrt{\gamma n}} \leq \sqrt{n} \frac{x-\gamma}{\sqrt{2\gamma}}).\\
\intertext{where $\mathbb{E}(C_1^2)=\frac{1}{n}$ and $\mathbb{V}(C_1^2) =\frac{\sqrt{2}}{n}$. Thus the Central Limit theorem holds and gives that the left hand side converges in distribution to a standard normal Gaussian when $n\to \infty$. The right hand side tends to $\text{sign}(x-\gamma)\infty$. Thus} 
F_{C_1^2 + \dots + C_{\gamma n}^2}(x) & \to \mathds{1}_{x\geq \gamma}(x).
\end{align*}
As we have a convergence in distribution towards a constant, the convergence in probability follows. 

\subsection{Distribution of hidden layers}\label{app:h_distribution}
Let us now consider at layer $l$ the weight matrix, $W^{(l)} \in \mathbb{R}^{N_{l}\times N_{l-1}}$, being of rank $r_l$: 
\begin{align*}W^{(l)} = \left[\begin{array}{c|c|c|c} 
& & & \\
\alpha^{(l)}_{1,1} C^{(l)}_1 + \dots + \alpha^{(l)}_{r_l,1} C^{(l)}_{r_l} & \alpha^{(l)}_{1,2} C^{(l)}_1 + \dots + \alpha^{(l)}_{r_l,2} C^{(l)}_{r_l} & \dots & \alpha^{(l)}_{1, N_{l-1}} C^{(l)}_1 + \alpha^{(l)}_{r_l,N_{l-1}} C^{(l)}_{r_l}\\
& & & 
\end{array}\right], \end{align*} where, at any layer $l$, for any $k \in \llbracket 1, r_l \rrbracket $, the scalars $\left(\alpha^{(l)}_{k,i}\right)_{1\leq i \leq N_{l-1}}\in \mathbb{R}$ are identically and independently drawn from a Gaussian distribution $\mathcal{N}(0,\frac{\sigma_{\alpha}^2}{N_{l-1}})$ and the columns $C^{(l)}_1, \dots, C^{(l)}_{r_l}$ are drawn jointly as the matrix $C^{(l)}:=[C^{(l)}_1, \dots, C^{(l)}_{r_l}]\in\mathbb{R}^{N_l\times r_l}$ from the Grassmannian of rank $r$ matrices with orthonormal columns having zero mean and variance $1/N_l$.  Similarly, in this section, we consider a random bias at layer $l$ along the directions given by $C^{(l)}_1, \dots, C^{(l)}_{r_l}$, i.e. a bias of the form $b^{(l)} (C^{(l)}_1 + \dots + C^{(l)}_{r_l})$, where $b^{(l)}\in \mathbb{R} \sim \mathcal{N}(0, \sigma_b^2)$. 

\vspace{0.2cm}
Thus, at layer $l$, whose width is $N_l$, the preactivation vector $h^{(l)} \in \mathbb{R}^{N_l}$ is given by
\begin{align*}
    h^{(l)} &= W^{(l)} \phi(h^{(l-1)}) + b^{(l)} (C_1^{(l)} + \dots + C_{r_l}^{(l)}) \\ 
    & = C_1^{(l)}\bigg(\underbrace{\sum\limits_{j=1}^{N_{l-1}} \alpha_{1,j}^{(l)} \phi(h^{(l-1)}_j) + b^{(l)}}_{:=z_1^{(l)}}\bigg) + C_2^{(l)}\bigg(\underbrace{\sum\limits_{j=1}^{N_{l-1}} \alpha_{2,j}^{(l)} \phi(h^{(l-1)}_j)+ b^{(l)} }_{:=z_2^{(l)}}\bigg) + \dots + C_{r_l}^{l} \bigg(\underbrace{\sum\limits_{j=1}^{N_{l-1}} \alpha_{r_l,j}^{(l)} \phi(h^{(l-1)}_j) + b^{(l)}}_{:=z_{r_l}^{(l)}}\bigg) \\
    & = \sum\limits_{k=1}^{r_{l}} \underbrace{z_k^{(l)} }_{\in \mathbb{R}} C_k^{(l)},
\end{align*}
where the scalars $z_k^{(l)}$ follow a Gaussian distribution, given the preactivation vector at the previous layer $z_k^{(l)}|_{h^{(l-1)}} \sim \mathcal{N}\bigg(0, \frac{\sigma^2_{\alpha}}{N_{l-1}} \sum\limits_{j=1}^{N_{l-1}} \phi(h^{(l-1)}_j)^2 + \sigma_b^2 \bigg)$, which is given using the Central Limit Theorem in the large width  $N_{l-1}$ regime.

\iffalse
\begin{itemize}
\item If $C_1^{(l)}, \dots, C_{r_l}^{(l)}$ are given, then $h^{(l)}|_{h^{(l-1)}}$ follows a Gaussian distribution, whose entries are not independent. 
\item If the column vectors are not given, $h^{(l)}|_{h^{(l-1)}}$ is a sum of $r_l$ $\chi^2$ distributions.
\end{itemize}
\fi

\subsection{Length recursion formula}\label{SM:length_recursion}
This being said, one can compute the length of the (random) preactivation vector, at layer $l$.
\begin{align*}
q^l & := \frac{1}{N_l} ||h^{(l)}||^2_2 = \frac{1}{N_l} \sum\limits_{j=1}^{N_l} (h_j^{(l)})^2 \\
& = \frac{1}{N_l} \left((z_1^{(l)})^2 ||C^{(l)}_1||_2^2 + \dots + (z_{r_l}^{(l)})^2 ||C^{(l)}_{r_l}||_2^2 \right) \quad \text{ using Pythagore's theorem} \\
& = \frac{1}{N_l} \left((z_1^{(l)} )^2 + \dots (z_{r_l}^{(l)})^2 \right) \quad \text{ since for any $k$, } ||C^{(l)}_k||_2 = 1\\
& = \frac{1}{N_l} \sum\limits_{k=1}^{r_l} (z_k^{(l)} )^2
\end{align*}
Therefore, given $h^{(l-1)}$, $q^l$ is a sum of $r_l$ $\chi^2$ distributions.

Let us now consider at any layer, a rank that is proportional to the width: $r_l = \gamma_l N_l$, where $\gamma_l \in (0, 1]$.
Thus,
\begin{align}
q^l = \gamma_l \frac{1}{\gamma_l N_l} \sum\limits_{k=1}^{\gamma_l N_l} (z_k^{(l)} )^2.
\end{align}
Then all $z_k^{(l)}$ are independent and identically distributed Gaussian variables. The Law of Large Numbers holds and $q^l \to \gamma_l \mathbb{V}\big(z_1^{(l)}\big)$ when $N_l \to \infty$, with
\begin{align*}
\mathbb{V}\big(z_1^{(l)}\big) = \frac{\sigma^2_{\alpha}}{N_{l-1}} \sum\limits_{j=1}^{N_{l-1}} \phi(h^{(l-1)}_j)^2 + \sigma^2_b.
\end{align*}

On the other hand, 
\begin{align*}
    h^{(l-1)}_j = \sum\limits_{k=1}^{r_{N_{l-1}}} z^{(l-1)}_k \big(C^{(l-1)}_k\big)_j = \big(z^{(l-1)}\big)^T \big(C^{(l-1)}_.\big)_j
    \intertext{denoting } \big(C^{(l)}_.\big)_j := \begin{pmatrix} \big(C^{(l)}_1\big)_j\\ 
    \vdots \\
    \big(C^{(l)}_{r_{l}}\big)_j
    \end{pmatrix}
\end{align*}

We know the distribution of $z^{(l-1)} \sim \mathcal{N}\bigg( \mathbb{O}_{\mathbb{R}^{r_{l-1}}}, \frac{q^{l-1}}{\gamma_{l-1}} \mathbb{I}_{r_{l-1}} \bigg)$ and so, given $C^{(l-1)}$, $h^{(l-1)}_j \sim \mathcal{N}\bigg(0, \big(C^{(l-1)}_.\big)^T_j \frac{q^{l-1}}{\gamma_{l-1}} \mathbb{I}_{r_{l-1}} \big(C^{(l-1)}_.\big)_j \bigg)$.

In the asymptotic limit approximation, when $N_{l-1}\to \infty$, then $\big(C^{(l-1)}_.\big)^T_j \big(C^{(l-1)}_.\big)_j \to \gamma_{l-1}$ in probability as shown in  \cref{lemma:convergence_norm_columns}. Therefore, $h^{(l-1)}_j \sim \mathcal{N}\big(0, q^{l-1} \big)$.

In the limit when $N_{l-1} \to \infty$, the Law of Large Numbers enables us to conclude,
\begin{align*}
q^l & = \gamma_l \big(\frac{\sigma^2_{\alpha}}{N_{l-1}} \sum\limits_{j=1}^{N_{l-1}} \phi(h^{(l-1)}_j)^2 + \sigma^2_b \big) \\
& \to \gamma_l \big( \sigma^2_{\alpha} \int_{\mathbb{R}} \phi^2(\sqrt{q^{l-1}}z) Dz + \sigma_b^2\big):= \mathcal{V}(q^{(l-1)}|\sigma^2_{\alpha}, \sigma^2_b, \gamma_l).
\end{align*}

Note that when $\forall l, \gamma_l = 1$ and $\sigma_{\alpha} = \sigma_W$, one recovers the formula from \cite{Poole_2016}. Alternatively, by rescaling the variances by $\gamma_l$ at every layer, e.g. $\sigma^2_{\alpha} \to \frac{\sigma^2_W}{\gamma_l}$ and $\sigma_b^2 \to \frac{\sigma^2_b}{\gamma_l}$, we recover the formulae of \cite{Poole_2016}.

\subsection{Correlation recursion formula}\label{SM:covariance_recursion}
Let us denote by $x^{0,1}$ and $x^{0,2}$ two input data. Then one can define the following 2 by 2 matrix
\begin{align*}
    (q^l_{ab})_{1 \leq a,b \leq 2} = \frac{1}{N_l} \sum\limits_{i=1}^{N_l} \begin{pmatrix}  h_i^{(l)}(x^{0,1})^2 & h_i^{(l)}(x^{0,1})h_i^{(l)}(x^{0,2})\\
    h_i^{(l)}(x^{0,1})h_i^{(l)}(x^{0,2}) & h_i^{(l)}(x^{0,2})^2
    \end{pmatrix}
\end{align*}
where, for $i \in \llbracket 1, N_l \rrbracket$, $h_i^{(l)}(x^{0,a})= \sum\limits_{r=1}^{r_{l}} z_k^{(l)}(x^{0,a}) \big(C^{(l)}_k\big)_i$.

So, 
\begin{align*}
\frac{1}{N_l} \sum\limits_{i=1}^{N_l} h_i^{(l)}(x^{0,1})h_i^{(l)}(x^{0,2}) &= \frac{1}{N_l} \sum\limits_{i=1}^{N_l} \bigg(\sum\limits_{k=1}^{r_{l}} z_k^{(l)}(x^{0,1}) \big(C^{(l)}_k\big)_i \bigg) \bigg( \sum\limits_{p=1}^{r_{l}} z_p^{(l)}(x^{0,2}) \big(C^{(l)}_p\big)_i \bigg) \\
& = \frac{1}{N_l} \sum\limits_{k=1}^{r_{l}} z_k^{(l)}(x^{0,1}) z_k^{(l)}(x^{0,2}) \bigg( \underbrace{\sum\limits_{i=1}^{N_l}  \big(C^{(l)}_k\big)_i^2 }_{||C^{(l)}_k||_2^2 = 1} \bigg) +  \frac{1}{N_l} \sum\limits_{\substack{1\leq k,p \leq r_l \\ k \neq p}} z_k^{(l)}(x^{0,1}) z_p^{(l)}(x^{0,2}) \bigg( \underbrace{\sum\limits_{i=1}^{N_l}  \big(C^{(l)}_k\big)_i \big(C^{(l)}_p\big)_i}_{\langle C^{(l)}_k, C^{(l)}_p \rangle = 0} \bigg) \\
& = \frac{1}{N_l} \sum\limits_{k=1}^{r_{l}} z_k^{(l)}(x^{0,1}) z_k^{(l)}(x^{0,2})\\
\intertext{Therefore, when the rank is proportional to the width and the width $N_l \to \infty$ as previously, the Law of Large Numbers gives}
& = \gamma_l \frac{1}{\gamma_l N_l} \sum\limits_{k=1}^{r_{l}} z_k^{(l)}(x^{0,1}) z_k^{(l)}(x^{0,2}) \to \gamma_l Cov\bigg(z_1^{(l)}(x^{0,2}), z_1^{(l)}(x^{0,2})\bigg)
\end{align*}

On the other hand, 
\begin{align*}
Cov\bigg(z_1^{(l)}(x^{0,1}), z_1^{(l)}(x^{0,2})\bigg) & = \sum\limits_{1\leq i,j \leq N_{l-1}} \phi(h^{(l-1)}_i(x^{0,1})) \phi(h^{(l-1)}_j(x^{0,2})) \underbrace{Cov(\alpha^{(l)}_{1,i}, \alpha^{(l)}_{1,j})}_{\frac{\sigma_{\alpha} }{N_{l-1}}\delta_{i,j}} \\ & + \sum\limits_{1\leq i\leq N_{l-1}} \phi(h^{(l-1)}_i(x^{0,1})) \underbrace{Cov(\alpha^{(l)}_{1,i}, b^{(l)}_1)}_{ = 0} \\ & +  \sum\limits_{1\leq i\leq N_{l-1}} \phi(h^{(l-1)}_i(x^{0,2})) \underbrace{Cov(\alpha^{(l)}_{1,i}, b^{(l)}_1)}_{ = 0} + \underbrace{Cov(b^{(l)}_1, b^{(l)}_1)}_{\sigma_b^2}
\end{align*}
Thus, when $N_{l-1}\to \infty$, the Law of Large Numbers enables us to conclude
\begin{align*}
q^l_{12} = \gamma_l \bigg( \sigma^2_{\alpha} \int_{\mathbb{R}^2} \phi(\sqrt{q^{l-1}_{11}}z_1)\phi(\sqrt{q^{l-1}_{22}}(c_{12}^{l-1} z_1 + \sqrt{1-(c_{12}^{l-1})^2}z_2) Dz_1 Dz_2 + \sigma^2_b \bigg)
\intertext{with $c_{12}^{l} = q_{12}^l (q_{11}^l q_{22}^l)^{-\frac{1}{2}}$.}
\end{align*}

Note that if we consider the short convergence of the variance, compared to the covariance one, as observed in \cite{Poole_2016}, then we can assume $q_{11}^l \approx q_{22}^{l} \approx q^*$. We can then rescale the previous covariance map to get the correlation map as follows:

\begin{align*}
c^l_{12} = \frac{\gamma_l}{q^*} \bigg( \sigma^2_{\alpha} \int_{\mathbb{R}^2} \phi(\sqrt{q^*}z_1)\phi(\sqrt{q^*}(c_{12}^{l-1} z_1 + \sqrt{1-(c_{12}^{l-1})^2}z_2) Dz_1 Dz_2 + \sigma^2_b \bigg)
\end{align*}

We observe that $1$ is always a fixed point as $1 = \frac{\gamma_l}{q^*} \bigg( \sigma^2_{\alpha} \int_{\mathbb{R}} Dz\phi^2(\sqrt{q^*}z) + \sigma_b^2 \bigg) = \frac{1}{q^*} \mathcal{V}(q^*|\sigma_{\alpha}^2, \sigma_b^2, \gamma) = \frac{q^*}{q^*}$.

\begin{figure}[h!]
    \vskip 0.2in
    \begin{center}
\centerline{\includegraphics[width=0.5\columnwidth]{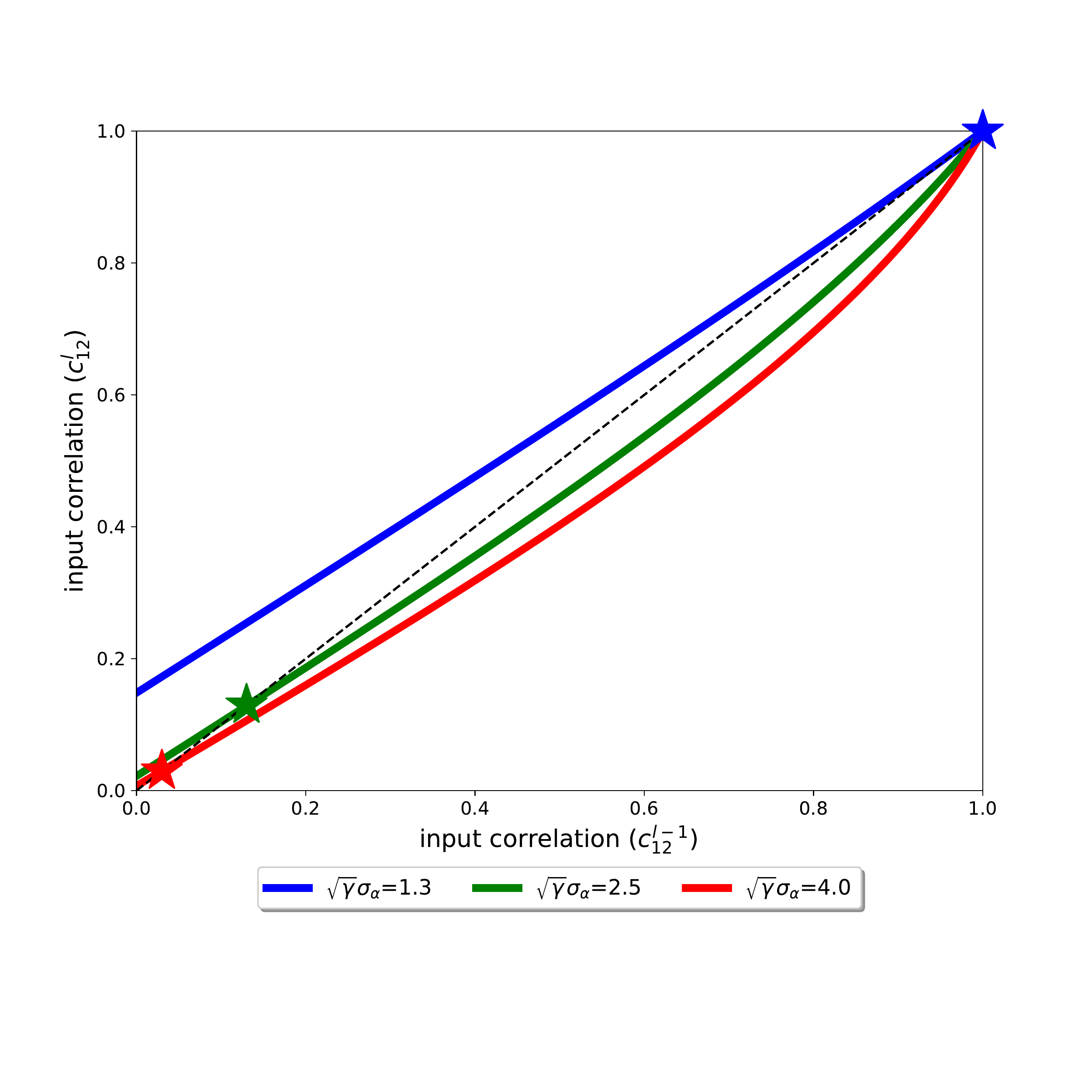}}
    \caption{Correlation map of a low-rank neural network where the rank is proportional to the width by a factor $\gamma$. The nonlinear activation function is $\phi = \text{tanh}$. The map is given by \eqref{eq:Cmap} and the integral is computed numerically. The dashed line is the identity function and stars represent fixed points of the correlation map.}
    \label{fig:correlation_map}
    \end{center}
    \vskip -0.2in
\end{figure}

\begin{figure}[h!]
    \vskip 0.2in
    \begin{center}
\centerline{\includegraphics[width=0.5\columnwidth]{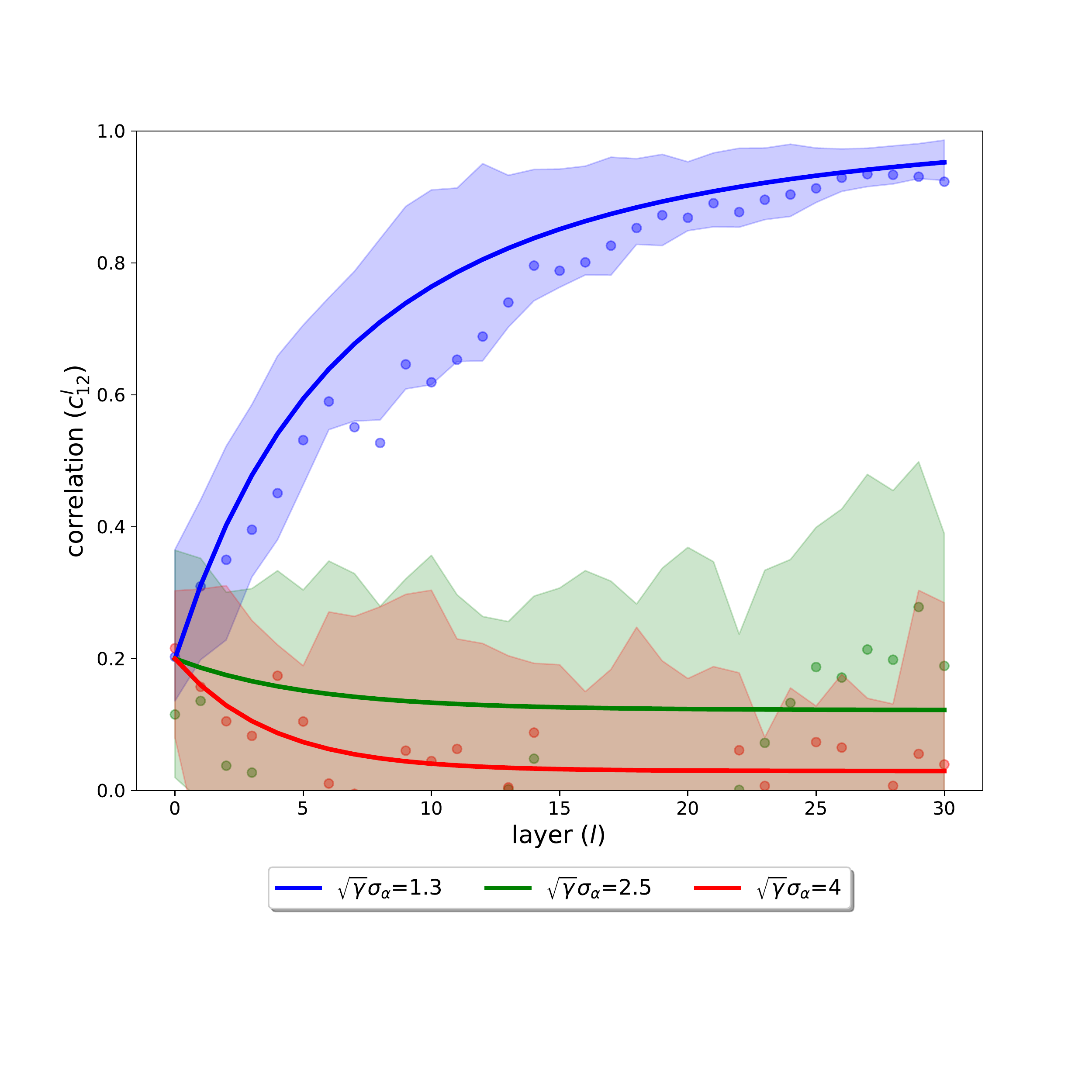}}
    \caption{Dynamic of the correlation through layers, starting from two input vectors with correlation $c^0_{12} = 0.2$. Points are obtained empirically and averaged over 5 simulations when the lines are derived from the theory, see \eqref{eq:Cmap}. Confidence intervals of 2 standard deviations around each point are shown. For each point on the plot, we generated a pair of points with correlation $c^0_{12}$, passed them through a Wide low-rank network initialised and computed the correlation between both preactivation vectors across layers. The network has constant width $N = 1000$. $\sigma_b = 0.3, \phi = \text{tanh}, \gamma=\frac{1}{4}$.}
    \label{fig:dynamics_c0_02}
    \end{center}
    \vskip -0.2in
\end{figure}

\begin{figure}[h!]
    \vskip 0.2in
    \begin{center}
\centerline{\includegraphics[width=0.5\columnwidth]{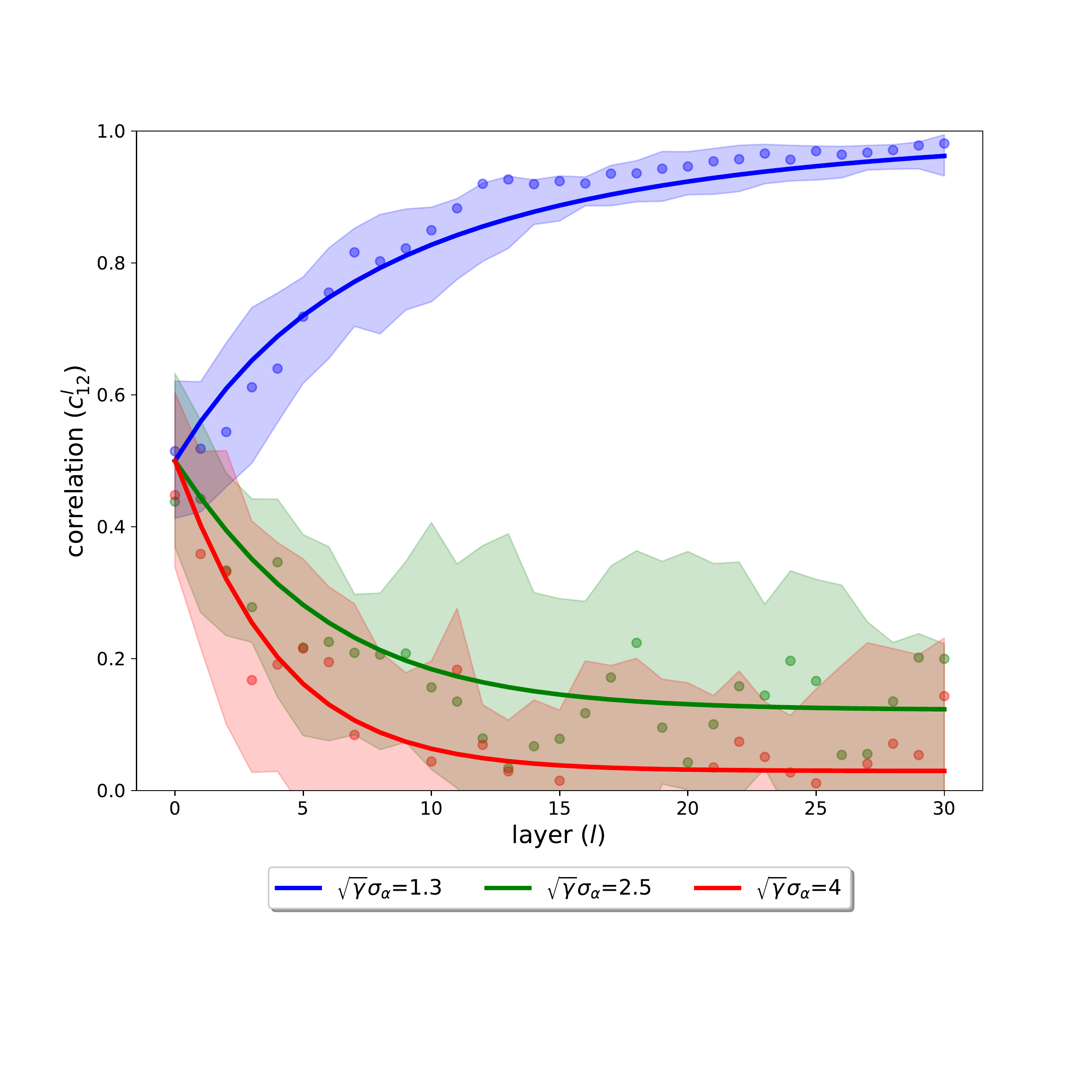}}
    \caption{Dynamic of the correlation through layers, starting from two input vectors with correlation $c^0_{12} = 0.5.$. Points are obtained empirically and averaged over 5 simulations when the lines are derived from the theory, see \eqref{eq:Cmap}. Confidence intervals of 2 standard deviations around each point are shown. For each point on the plot, we generated a pair of points with correlation $c^0_{12}$, passed them through a Wide low-rank network initialised and computed the correlation between both preactivation vectors across layers. The network has constant width $N = 1000$. $\sigma_b = 0.3, \phi = \text{tanh}, \gamma=\frac{1}{4}$.}
    \label{fig:dynamics_c0_05}
    \end{center}
    \vskip -0.2in
\end{figure}

\begin{figure}[h!]
    \vskip 0.2in
    \begin{center}
\centerline{\includegraphics[width=0.5\columnwidth]{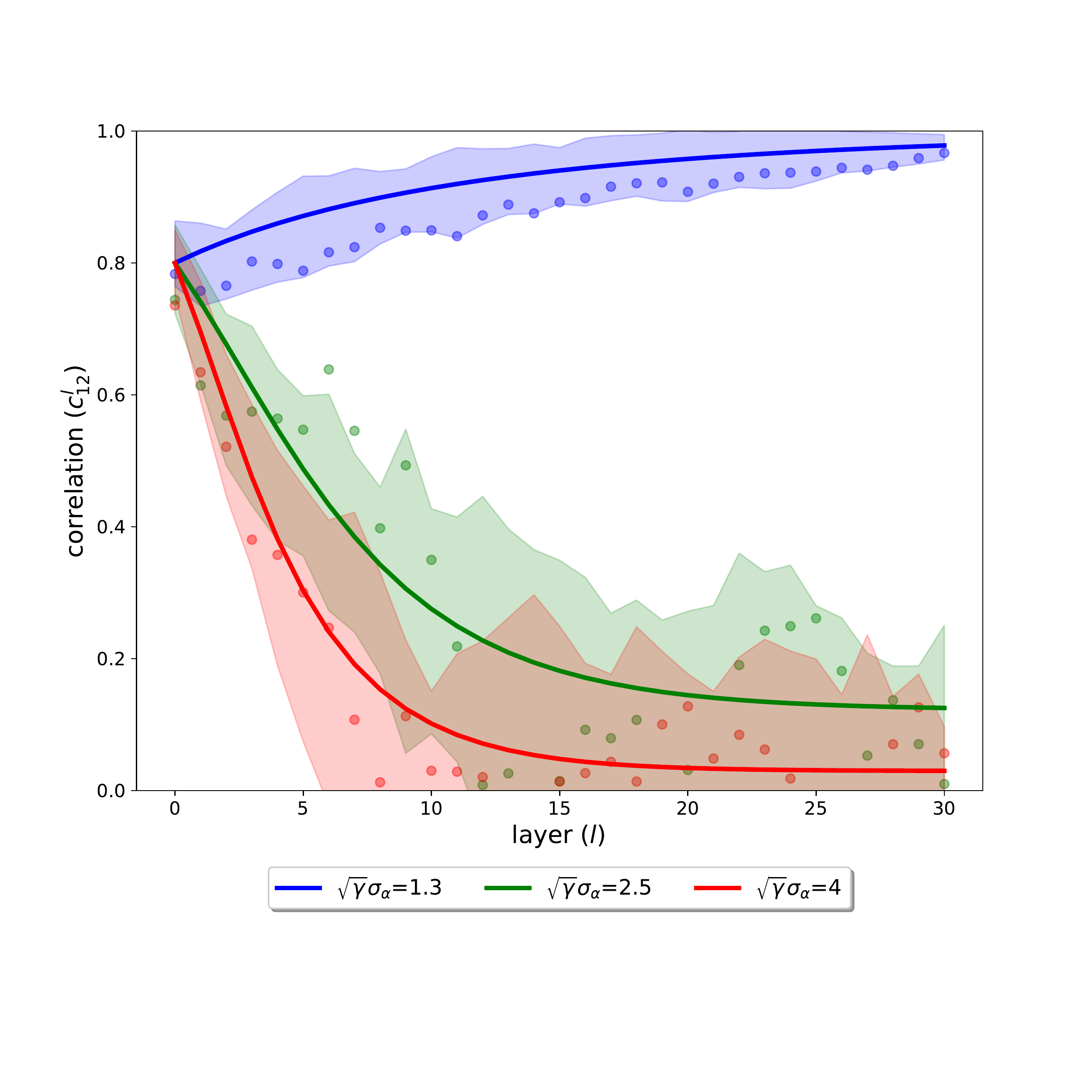}}
    \caption{Dynamic of the correlation through layers, starting from two input vectors with correlation $c^0_{12} = 0.8.$. Points are obtained empirically and averaged over 5 simulations when the lines are derived from the theory, see \eqref{eq:Cmap}. Confidence intervals of 2 standard deviations around each point are shown. For each point on the plot, we generated a pair of points with correlation $c^0_{12}$, passed them through a Wide low-rank network initialised and computed the correlation between both preactivation vectors across layers. The network has constant width $N = 1000$. $\sigma_b = 0.3, \phi = \text{tanh}, \gamma=\frac{1}{4}$.}
    \label{fig:dynamics_c0_08}
    \end{center}
    \vskip -0.2in
\end{figure}

For completeness we replicate correlation maps and dynamics of correlations through layers using the same parameters as in \cite{Poole_2016} suitably modified for the low-rank setting, see \cref{fig:correlation_map} - \ref{fig:dynamics_c0_08}.  The dynamics of the low-rank networks are observed to be consistent, under appropriate scaling by $\gamma_l$,with those of the full-rank networks in \cite{Poole_2016}.

\subsection{Derivative of the correlation map}\label{SM:jacobian_correlation}

In this section, we extend the computations of the derivative of the correlation map. 
\begin{align*}
\frac{\partial c^l_{12}}{\partial c^{l-1}_{12}} & =  \frac{\gamma_l}{q^*} \bigg( \sigma^2_{\alpha} \int_{\mathbb{R}^2} \phi(u_1)\phi^\prime(u_2) \big(\sqrt{q^*} z_1 - \sqrt{q^*} \frac{c^{l-1}_{12}}{\sqrt{1 - (c^{l-1}_{12})^2}} z_2\big) Dz_1 Dz_2 \bigg)
\intertext{Using, for $F$ smooth enough, the identity $\int_{\mathbb{R}} F(z)z Dz = \int_{\mathbb{R}} F^\prime(z) Dz$ to the functions $G:z_1\mapsto \phi(\sqrt{q^*}z_1) \int_{z_2} \phi^\prime\big(\sqrt{q^*}(c^{l-1}_{12} z_1 + \sqrt{1 - (c^{l-1}_{12})^2}z_2)\big)Dz_2$ and $H:z_2\mapsto \int_{z_1} \phi(\sqrt{q^*}z_1) \phi^\prime\big(\sqrt{q^*}(c^{l-1}_{12} z_1 + \sqrt{1 - (c^{l-1}_{12})^2}z_2)\big)Dz_1$, we obtain}
\frac{\partial c^l_{12}}{\partial c^{l-1}_{12}} & =  \gamma_l \sigma^2_{\alpha} \int_{\mathbb{R}^2} \phi^\prime(\sqrt{q^*}z_1)\phi^\prime\big(\sqrt{q^*}(c_{12}^{l-1} z_1 + \sqrt{1-(c_{12}^{l-1})^2}z_2)\big) Dz_1 Dz_2,
\intertext{which, evaluated at its fixed point $c^{l-1}_{12} = 1$ gives }
\chi_{\gamma} &:= \frac{\partial c^l_{12}}{\partial c^{l-1}_{12}}|_{c^{l-1}_{12} = 1} = \gamma_l \sigma^2_{\alpha} \int_{\mathbb{R}} \bigg(\phi^\prime(\sqrt{q^*}z)\bigg)^2 Dz.
\end{align*}

\begin{figure}[ht]
    \vskip 0.2in
    \begin{center}
    \centerline{\includegraphics[width=0.5\columnwidth]{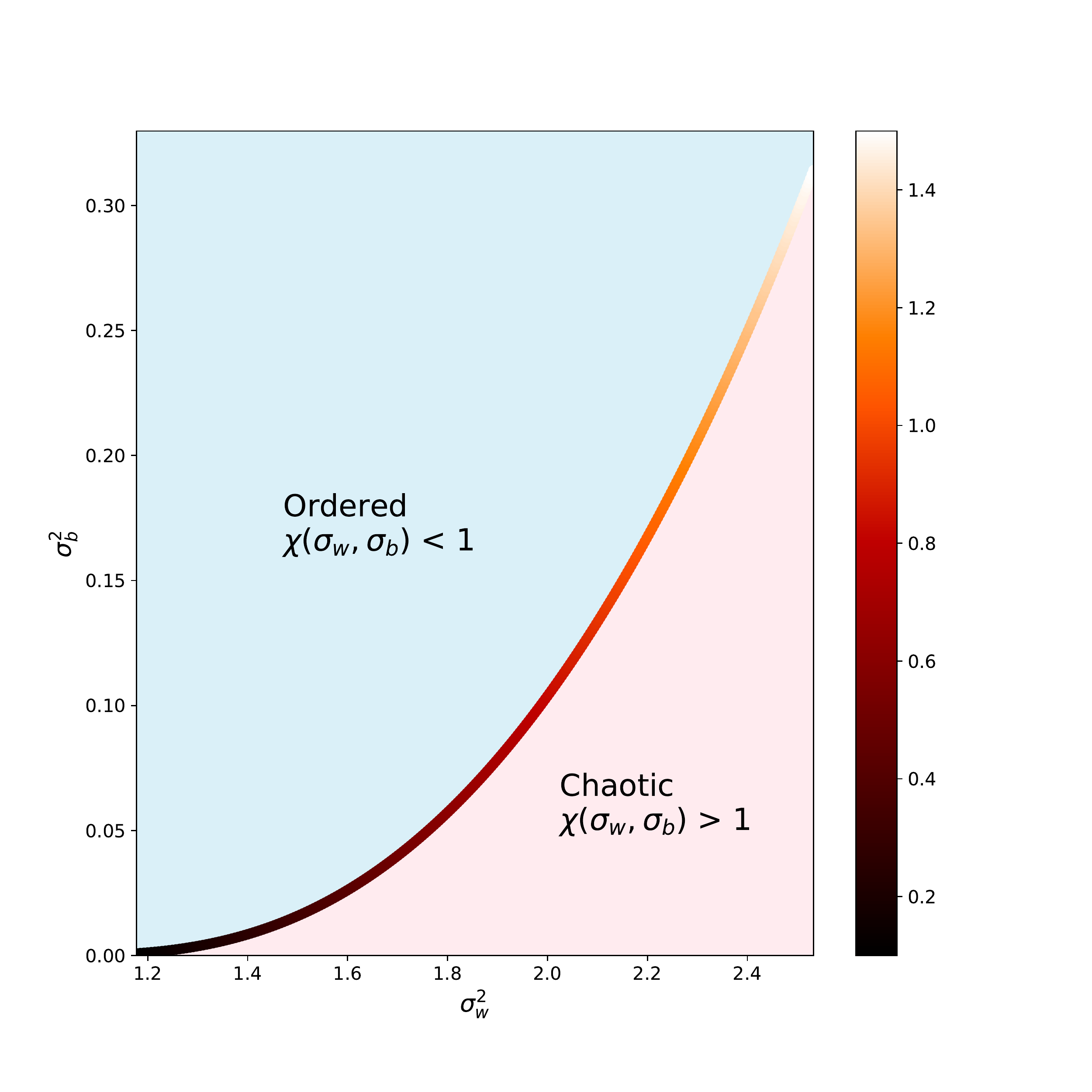}}
    \caption{Original edge-of-chaos curve of the full-rank feedforward neural network with nonlinear activation $\phi(x)=tanh(x)$.}
    \label{fig:EOC_gamma1}
    \end{center}
    \vskip -0.2in
\end{figure}

The edge-of-chaos level set defined by $\chi_{\gamma}=1$ for nonlinear activation $\phi(x)=tanh(x)$ is shown in Figure \ref{fig:EOC_gamma14} with axes $\gamma\sigma_w^2$ and $\gamma\sigma_b^2$.
Figure \ref{fig:EOC_gamma1} show the analogous edge-of-chaos plot for a full-rank matrix as given, which is identical to that of \ref{fig:EOC_gamma14} but with axes $\sigma_w^2$ and $\sigma_b^2$.

\subsection{Length depth scale}\label{SM:length_scale}

Recall that $q^l = \gamma_l \big(\sigma_{\alpha}^2 \int Dz \phi^2(\sqrt{q^{l-1}}z) + \sigma_b^2 \big)$ and $q^*$ is a fixed point assumed to exist when $\gamma_l = \gamma$ at any layer $l$. We then define the perturbation $\epsilon_l \to 0$ such that $q^l = q^* + \epsilon_l$. We can then expand the relation around its fixed point, as done in the case of feedforward neural network in \cite{Schoenholz_2017}.
\begin{align*}
q^{l+1} & = q^* + \epsilon_{l+1} = \gamma \big(\sigma_{\alpha}^2 \int Dz \phi^2((\epsilon_l + q^{*})^{\frac{1}{2}} z) + \sigma_b^2 \big) \\
& = \gamma \bigg(\sigma_{\alpha}^2 \int Dz \phi\big(\sqrt{q^*}z + \frac{1}{2} \frac{\epsilon_l}{\sqrt{q^*}} z + \mathcal{O}(\epsilon_l^2)\big)^2 + \sigma_b^2\bigg) \text{ expanding the square root}\\
& = \gamma \bigg(\sigma_{\alpha}^2 \int Dz \big((\phi(\sqrt{q^*}z) + \phi^\prime(\sqrt{q^*}z) \frac{\epsilon_l}{2\sqrt{q^*}} z + \mathcal{O}(\epsilon_l^2)\big)^2 + \sigma_b^2\bigg) \text{ expanding $\phi$ around $\sqrt{q^*}z$} \\
& = \gamma \bigg(\sigma_{\alpha}^2 \int Dz \phi^2(\sqrt{q^*}z) + \int Dz \phi^\prime(\sqrt{q^*}z)\phi(\sqrt{q^*}z) \frac{\epsilon_l}{\sqrt{q^*}} z + \mathcal{O}(\epsilon_l^2) + \sigma_b^2\bigg) \\
& = q^* + \gamma \int Dz \phi^\prime(\sqrt{q^*}z)\phi(\sqrt{q^*}z) \frac{\epsilon_l}{\sqrt{q^*}} z + \mathcal{O}(\epsilon_l^2) \text{ by definition of $q^*$} \\
& = q^* + \epsilon_l \gamma \sigma_{\alpha}^2 \bigg( \int Dz \big(\phi^\prime(\sqrt{q^*}z)\big)^2 + \int Dz \phi^{\prime\prime}(\sqrt{q^*}z)\phi(\sqrt{q^*}z) \bigg) + \mathcal{O}(\epsilon_l^2) \text{ using $\int DzF(z)z = \int Dz F^\prime(z)$} \\
\end{align*}
Note that in the proof above we assumed the activation function $\phi$ to be smooth enough to use its Taylor expansion around the point $\sqrt{q^*}z$ for any $z$.

Therefore by identification, $\epsilon_{l+1} = \epsilon_l \big(\chi_{\gamma} + \gamma \sigma^2_{\alpha} \int Dz \phi^{\prime\prime}(\sqrt{q^*}z)\phi(\sqrt{q^*}z) \big) + \mathcal{O}(\epsilon_l^2)$, which concludes the proof.

\subsection{Correlation depth scale}\label{SM:correlation_scale}

Let consider the computation is done at a layer $l$ deep enough so that the variance map has already converged towards its fixed point $q^l_{11} = q^l_{22} = q^*$. We generate a perturbation $\epsilon_l \underset{l\to \infty}{\to} 0$ around the fixed point $c^*$ and analyse how it propagates: $c^{l}_{12} = c^* + \epsilon_l$.
Additionally, we introduce $u_1^l = \sqrt{q^*} z = u_1^*, u_2^* = \sqrt{q^*}(c^* z_1 + \sqrt{1-(c^*)^2} z_2)$ and $u_2^l = \sqrt{q^*} (c^l_{12} z_1 + \sqrt{1-(c^l_{12})^2} z_2)$. Following the same strategy as in the previous section (using expansions), it is shown in \cite{Schoenholz_2017} that 
\begin{align*}
    u_2^l = \begin{cases} u_2^* + \sqrt{q^*}\epsilon_l \big(z_1 - \frac{c^*}{\sqrt{1 - {c^*}^2}} z_2 \big) + \mathcal{O}(\epsilon_l^2) & \text{ when $c^* < 1$,}\\
    u_2^* + \sqrt{2q^*\epsilon_l}z_2 - \epsilon_l\sqrt{q^*}z_1 + \mathcal{O}(\epsilon_l^{\frac{3}{2}}) & \text{ when $c^* = 1$}.
    \end{cases}
\end{align*}

Therefore, in the first case $c^*<1$, 
\begin{align*}
c^{l+1}_{12} & = c^* + \epsilon_l = \frac{\gamma_l}{q^*} \bigg( \sigma^2_{\alpha} \int_{\mathbb{R}^2}Dz_1 Dz_2 \phi(u^*_1)\phi(u_2^l) + \sigma^2_b \bigg) \\
& = \frac{\gamma_l}{q^*} \bigg( \sigma^2_{\alpha} \int_{\mathbb{R}^2} Dz_1 Dz_2 \phi(u^*_1)\phi( u_2^* + \sqrt{q^*}\epsilon_l \big(z_1 - \frac{c^*}{\sqrt{1 - {c^*}^2}} z_2 \big) + \mathcal{O}(\epsilon_l^2)) + \sigma^2_b \bigg) \\
& = \frac{\gamma_l}{q^*} \bigg( \sigma^2_{\alpha} \int_{\mathbb{R}^2} Dz_1 Dz_2\phi(u^*_1)[\phi(u_2^*) + \phi^\prime (u_2^*)\sqrt{q^*}\epsilon_l \big(z_1 - \frac{c^*}{\sqrt{1 - {c^*}^2}} z_2 \big)] + \mathcal{O}(\epsilon_l^2)  + \sigma^2_b \bigg) \text{ expanding $\phi$ around $u_2^*$} \\
& = c^* + \frac{\gamma_l}{\sqrt{q^*}} \sigma_{\alpha}^2 \epsilon_l \int_{\mathbb{R}^2} Dz_1 Dz_2\phi(u^*_1) \phi^\prime (u_2^*) z_1 - \frac{c^*}{\sqrt{1 - {c^*}^2}} \frac{\gamma_l}{\sqrt{q^*}} \int_{\mathbb{R}^2} Dz_1 Dz_2\phi(u^*_1) \phi^\prime (u_2^*) z_2 +\mathcal{O}(\epsilon_l^2) \\
& = c^* + \gamma_l \sigma_{\alpha}^2 \epsilon_l \int_{\mathbb{R}^2} Dz_1 Dz_2\phi(u^*_1) \big(\phi^\prime (u_2^*) + c^* \phi^{\prime\prime}(u_2^*)\big) - c^*\gamma_l \int_{\mathbb{R}^2} Dz_1 Dz_2\phi(u^*_1) \phi^{\prime}\prime (u_2^*) +\mathcal{O}(\epsilon_l^2)  \\
& = c^* + \gamma_l \sigma_{\alpha}^2 \epsilon_l \int_{\mathbb{R}^2} Dz_1 Dz_2\phi(u^*_1) \phi^\prime (u_2^*) + \mathcal{O}(\epsilon_l^2)
\end{align*}
where the second to last line is obtained using $\int DzF(z)z = \int Dz F^\prime(z)$, for $\phi$ smooth enough. We can then identify $\epsilon_{l+1} = \epsilon_l \gamma_l \sigma_{\alpha}^2 \int_{\mathbb{R}^2} Dz_1 Dz_2\phi(u^*_1) \phi^\prime (u_2^*) + \mathcal{O}(\epsilon_l^2) $.

In the second case $c^*=1$, $u_1^* = u_2^*$, $c^l_{12} = 1-\epsilon_l$ and we expand $\phi$ around $u_2^*$ until to the second order.
\begin{align*}
c^{l+1}_{12} & = 1 - \epsilon_{l+1} = \frac{\gamma_l}{q^*} \bigg( \sigma^2_{\alpha} \int_{\mathbb{R}^2}Dz_1 Dz_2 \phi(u^*_1)\phi(u_2^l) + \sigma^2_b \bigg) \\
& = \frac{\gamma_l}{q^*} \bigg( \sigma^2_{\alpha} \int_{\mathbb{R}^2} Dz_1 Dz_2 \phi(u^*_1)\phi\big( u_2^* + \sqrt{q^*}\epsilon_l \big(z_1 - \frac{c^*}{\sqrt{1 - {c^*}^2}} z_2 \big) + \mathcal{O}(\epsilon_l^2)\big) + \sigma^2_b \bigg) \\
& = \frac{\gamma_l}{q^*} \bigg( \sigma^2_{\alpha} \int_{\mathbb{R}^2} Dz_1 Dz_2 \phi(u^*_1)\bigg(\phi( u_2^*) + \phi^\prime( u_2^*)\big(\sqrt{2q^*\epsilon_l} z_2 - \sqrt{q^*}\epsilon_l z_1\big) + \phi^{\prime \prime}(u_2^*) \frac{1}{2} \big(\sqrt{2q^*\epsilon_l} z_2 - \sqrt{q^*}\epsilon_l z_1\big)^2 + \mathcal{O}(\epsilon_l^{\frac{3}{2}})\big) + \sigma^2_b \bigg) \\
& = c^* + \frac{\gamma_l}{\sqrt{q^*}} \sqrt{2\epsilon_l} \sigma^2_{\alpha} \int_{\mathbb{R}} Dz_1 \phi(u^*_1) \phi^\prime(u_2^*)\underbrace{\int_{\mathbb{R}}z_2Dz_2}_{= 0} -  \frac{\gamma_l}{\sqrt{q^*}} \epsilon_l \sigma^2_{\alpha} \int_{\mathbb{R}} Dz_1 \phi(u^*_1) \phi^\prime(u_2^*)z_1 \underbrace{\int_{\mathbb{R}} Dz_2}_{= 1}\\& \qquad \qquad+ \gamma_l \epsilon_l \sigma^2_{\alpha} \int_{\mathbb{R}} Dz_1  \phi(u^*_1) \phi^{\prime\prime}(u_2^*) \underbrace{\int_{\mathbb{R}}z_2^2 Dz_2}_{= 1} + \mathcal{O}(\epsilon_l^{\frac{3}{2}})\\
& = c^* - \frac{\gamma_l}{\sqrt{q^*}} \epsilon_l \sigma^2_{\alpha} \int_{\mathbb{R}} Dz_1 \phi(u^*_1) \phi^\prime(u_2^*)z_1 + \gamma_l \epsilon_l \sigma^2_{\alpha} \int_{\mathbb{R}} Dz_1  \phi(u^*_1) \phi^{\prime\prime}(u_2^*) + \mathcal{O}(\epsilon_l^{\frac{3}{2}})\\
& = c^* - \gamma_l \epsilon_l \sigma^2_{\alpha} \int_{\mathbb{R}} Dz_1  (\phi^\prime(u_1^*))^2 - \gamma_l \epsilon_l \sigma^2_{\alpha} \int_{\mathbb{R}} Dz_1 \phi(u_1^*) \phi^{\prime\prime}(u_2^*)+ \gamma_l \epsilon_l \sigma^2_{\alpha} \int_{\mathbb{R}} Dz_1  \phi(u^*_1) \phi^{\prime\prime}(u_2^*) + \mathcal{O}(\epsilon_l^{\frac{3}{2}})\\
& = c^* - \epsilon_l \gamma_l \sigma^2_{\alpha} \int_{\mathbb{R}} Dz  (\phi^\prime(\sqrt{q^*}z))^2 + \mathcal{O}(\epsilon_l^{\frac{3}{2}})
\end{align*}
Therefore, $\epsilon_{l+1} = \epsilon_l \gamma_l \sigma^2_{\alpha} \int_{\mathbb{R}} Dz  (\phi^\prime(\sqrt{q^*}z))^2 + \mathcal{O}(\epsilon_l^{\frac{3}{2}})$, which concludes the proof. 

\begin{figure}[ht]
    \vskip 0.2in
    \begin{center}
\centerline{\includegraphics[width=0.5\columnwidth]{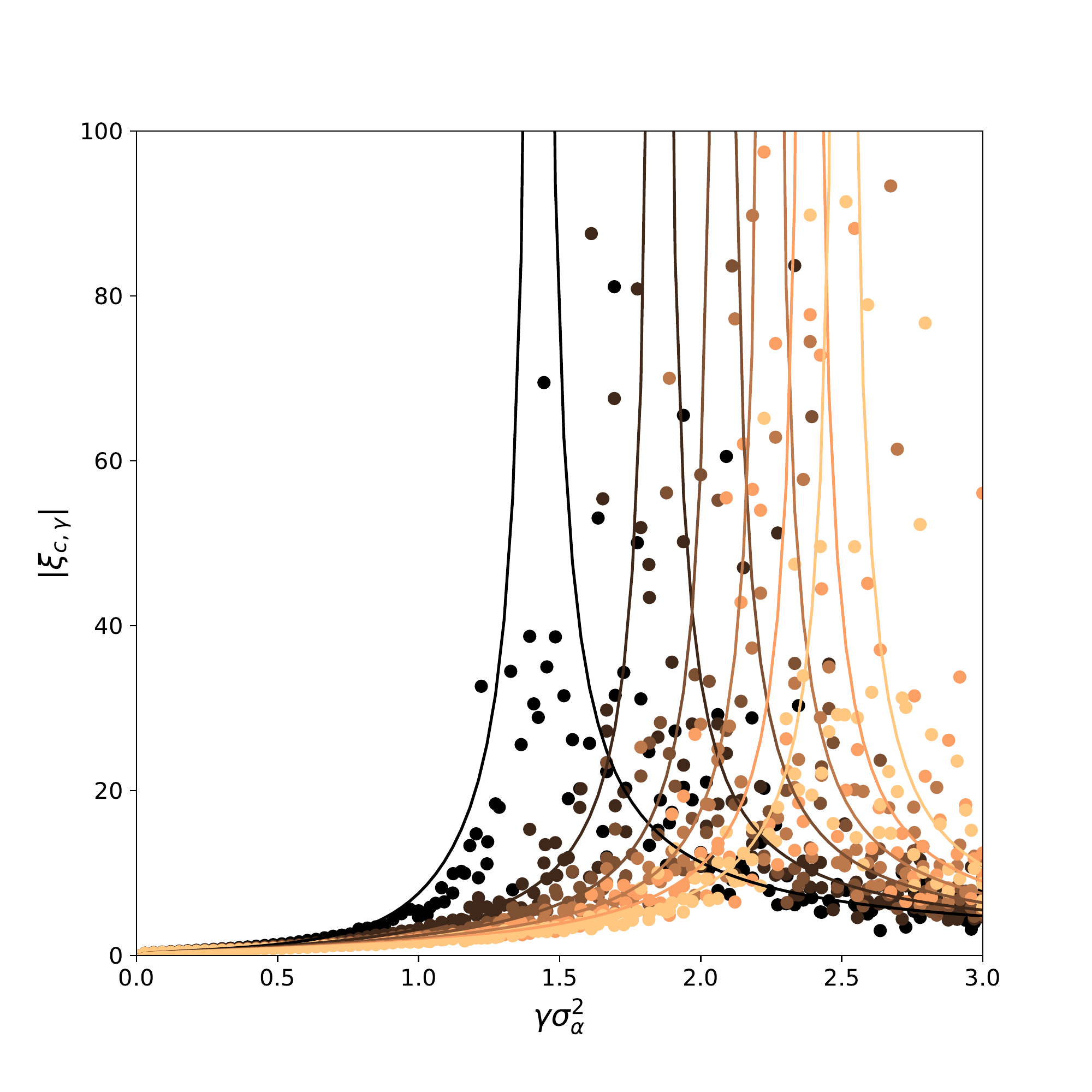}}
    \caption{Correlation depth scale with respect to $\gamma \sigma^2_{\alpha}$ diverging when $\chi_{\gamma}=1$. Points are obtained empirically when the lines are derived from the theory. The variance of the bias varies from $\sigma^2_b = 0.01 \gamma^{-1}$ (black) to $\sigma^2_b = 0.3 \gamma^{-1}$ (yellow). The network has constant width $N = 1000$. $\phi = \text{tanh}, \gamma=\frac{1}{4}$.}
    \label{fig:corr_depth_scale}
    \end{center}
    \vskip -0.2in
\end{figure}

\cref{fig:corr_depth_scale} shows the analytically calculated correlation depth scales as a function of $\gamma\sigma_{\alpha}^2$ as well as simulations with networks of width $N=1000$ and nonlinear activation $\phi(x)=tanh(x)$.  The networks depth scale are observed to be consistent with the analytic calculations; in particular showing the depth scale asymptotes at $\chi_{\gamma}=1$ for the different choices of $\sigma_b^2$.

\subsection{Backpropagation}\label{SM:backpropagation}

Recall that $h_i^{(l)} = \sum\limits_{k=1}^{r_{l}} \bigg(\sum\limits_{j=1}^{N_{l-1}} \alpha_{k,j}^{(l)} \phi(h^{(l-1)}_j) + b^{(l)} \bigg)(C_k^{(l)})_i$. Then the chain rule immediately gives,
\begin{align*}
\delta_j^l := \frac{\partial E}{\partial h_j^{(l)}} =  \big(\sum_{k=1}^{N_{l+1}} \delta_k^{l+1} W_{kj}^{(l+1)} \big) \phi^\prime(h_j^{(l)})
\end{align*}

Because the trainable parameters of our network are the coefficients $\alpha_{ij}^{(l)}$, we compute the gradient of the error loss $E$ with respect to them. So we need to adapt the proof from \cite{Schoenholz_2017}, derived in the standard feedforward case as follows. 

\begin{align*}
||\nabla_{\alpha_{ij}^{(l)}} E||^2_2 & = \sum_{i,j} \big( \frac{\partial E}{\partial \alpha_{ij}^{(l)}} \big)^2 \\
& \underset{N_l, N_{l+1} \to \infty}{\approx} N_l N_{l+1} \mathbb{E}\bigg(\big( \frac{\partial E}{\partial \alpha_{ij}^{(l)}} \big)^2\bigg).
\end{align*}
Since, assuming all these partial derivatives to be identically and independently distributed, the Law of Large numbers holds. 

On the one hand, 
\begin{align*}
\frac{\partial E}{\partial \alpha^{(l)}} & = \sum_{m=1}^{N_l} \frac{\partial E}{\partial h_m^{(l)}} \frac{\partial h_m^{(l)}}{\partial \alpha_{ij}^{(l)}} 
=  \sum_{m=1}^{N_l}  \delta_m^l \phi(h_j^{(l-1)}) (C_i^{(l)})_m = \bigg(\sum_{m=1}^{N_l} \delta_m^{l} (C_i^{(l)})_m \bigg) \phi(h_j^{(l-1)}).\\
\intertext{Therefore, assuming independence between the weights used for the forward pass and the weights backpropagated,}
\mathbb{E}\bigg(\big( \frac{\partial E}{\partial \alpha_{ij}^{(l)}} \big)^2\bigg) & = \mathbb{E}\bigg( \big(\sum_{m=1}^{N_l} \delta_m^{l} (C_i^{(l)})_m \big)^2 \bigg) \mathbb{E}\big(\phi^2(h_j^{(l-1)})\big) \\
& = \mathbb{E}\bigg( \sum_{m=1}^{N_l} (\delta_m^{l})^2 (C_i^{(l)})_m^2 + \sum_{p,m} \delta_m^{l} (C_i^{(l)})_m \bigg) \mathbb{E}\big(\phi^2(h_j^{(l-1)})\big) \\
& = \frac{1}{N_l} \sum_{m=1}^{N_l} \mathbb{E}\big((\delta_m^{l})^2\big) \mathbb{E}\big(\phi^2(h_j^{(l-1)})\big) \text{ since $(C_1^{(l)})_m \overset{\text{iid}}{\sim} \mathcal{N}(0, \frac{1}{N_l})$} \\
& = \mathbb{E}\big((\delta_1^{l})^2\big) \mathbb{E}\big(\phi^2(h_j^{(l-1)})\big).
\end{align*}
Therefore, the length of the gradient loss is proportional to the variance of $\delta_1^l$.

On the other hand, denoting with a subscript the function $f_a$ when fed with input data $x^{0,a}$,
\begin{align*}
\tilde{q}^l_{aa} &:= \mathbb{E}\big((\delta_{1,a}^{l})^2\big) \\
& = \mathbb{E}\bigg( \big(\sum_{k=1}^{N_{l+1}} \delta_{k,a}^{l+1} W_{kj}^{(l+1)} \big)^2 \bigg) \mathbb{E} \bigg(\big(\phi^\prime(h_{j,a}^{(l)})\big)^2\bigg)\\
& = \bigg( \sum_{k=1}^{N_{l+1}}  \mathbb{E}\big((\delta_{k,a}^{l+1})^2\big) \mathbb{E} \big((W_{kj}^{(l+1)})^2\big)\bigg) \mathbb{E} \bigg(\big(\phi^\prime(h_{j,a}^{(l)})\big)^2\bigg)
\intertext{where we used again the assumed independence. The first and second order moments of $W_{ij}^{(l)}$ are given by $\mathbb{E}(W_{ij}^{(l)}) = 0$ and $\mathbb{E}\big((W_{ij}^{(l)})^2\big) = \mathbb{E}\big((\sum\limits_{p=1}^{r_l} \alpha_{p,j}^{(l)} (C_p^{(l)})_i)^2\big) = \sum\limits_{p=1}^{r_l} \mathbb{V}(\alpha_{p,j}^{(l)}) \mathbb{V}((C_p^{(l)})_i) = r_l \frac{\sigma^2_{\alpha}}{N_{l-1}} \frac{1}{N_l} $. Thus,}
\tilde{q}^l_{aa} & =  r_{l+1} \frac{\sigma^2_{\alpha}}{N_{l}} \frac{1}{N_{l+1}} \bigg(\sum_{k=1}^{N_{l+1}}  \mathbb{E}\big((\delta_{k,a}^{l+1})^2\big) \bigg) \mathbb{E} \bigg(\big(\phi^\prime(h_{j,a}^{(l)})\big)^2\bigg) \\
& =  r_{l+1} \frac{\sigma^2_{\alpha}}{N_{l}} \frac{1}{N_{l+1}} N_{l+1} \tilde{q}^{l+1}_{aa} \mathbb{E} \bigg(\big(\phi^\prime(h_{j,a}^{(l)})\big)^2\bigg)\\
& = r_{l+1} \frac{\sigma^2_{\alpha}}{N_{l}} \frac{1}{N_{l+1}} N_{l+1} \tilde{q}^{l+1}_{aa} \int Dz \big(\phi^\prime(\sqrt{q_{aa}^{l-1}}z)\big)^2 \text{ as $h_{j,a}^{(l-1)} \sim \mathcal{N}(0, q_{aa}^{l-1})$.}
\intertext{Considering that the computation is done at a layer deep enough, since $q^{l-1}$ converges to $ q^*$ shortly, then $q_{aa}^{l-1}\approx q^*$, and, as $r_l = \gamma_l N_l$,}
\tilde{q}^l_{aa} & = \tilde{q}^{l+1}_{aa} \gamma_{l+1} \sigma^2_{\alpha} \int Dz \big(\phi^\prime(\sqrt{q^{*}}z)\big)^2 = \tilde{q}^{l+1}_{aa} \chi_{l+1}.
\end{align*}

\begin{figure}[ht]
    \vskip 0.2in
    \begin{center}
\centerline{\includegraphics[width=0.5\columnwidth]{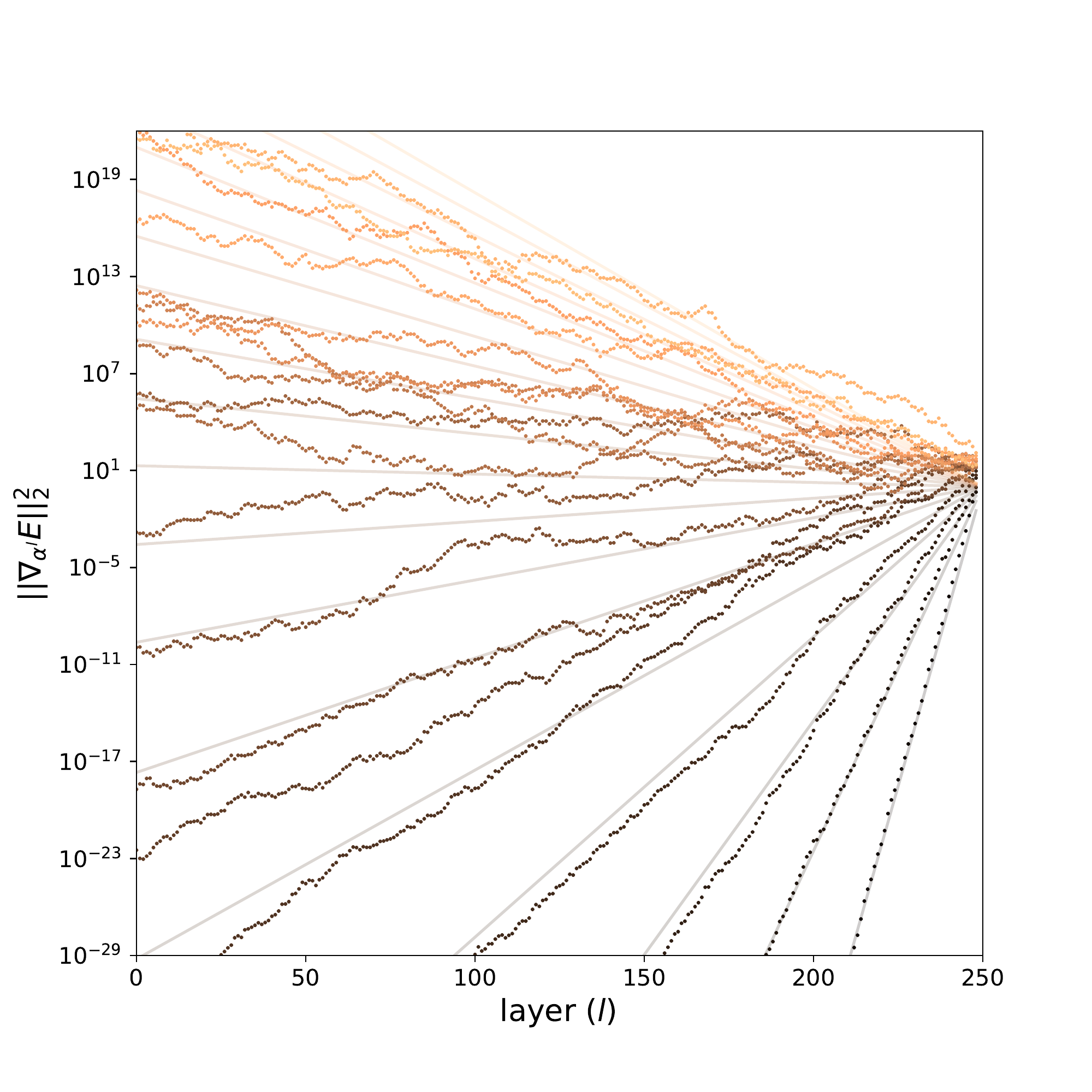}}
    \caption{Exponential evolution of the propagation of the $L_2$-norm of the gradient  with respect to the depth for a $250$ layer deep random neural network with a cross entropy loss on MNIST dataset. Points are obtained empirically when the lines are derived from the theory. The variance of the weights $\gamma \sigma^2_{\alpha}$ varies from $0.01\gamma^{-1}$ (black) to $ 0.3 \gamma^{-1}$ (yellow) when the variance of the bias $\gamma \sigma_b^2$ is kept fixed to $0.05$. $\phi = \text{tanh}, \gamma=\frac{1}{4}$.}
    \label{fig:gradient_propagation}
    \end{center}
    \vskip -0.2in
\end{figure}

\cref{fig:gradient_propagation} demonstrates the exponential evolution of $||\nabla_{\alpha^{(l)}} E||^2_2$ from the final layer, $L=250$, to the earlier layers.  The analytic expressions are shown to be consistent with simulation from random low-rank networks with nonlinear activation $\phi(x)=tanh(x)$, rank to width scale $\gamma=1/4$, the bias variance $\sigma_b^2$ held fixed and $\sigma_{\alpha}^2$ varying.

\subsection{Average singular value of $D^lW^{(l)}$}\label{SM:average_singular_value}

As a preliminary, let us first note that $\frac{1}{N_l} \text{Tr}\bigg(D^l W^{(l)}(D^l W^{(l)})^T\bigg) = \frac{1}{N_l} \sum\limits_{k=1}^{N_l} \lambda_k\bigg(D^l W^{(l)}(D^l W^{(l)})^T\bigg) = \frac{1}{N_l} \sum\limits_{k=1}^{N_l} \sigma^2_k\bigg(D^l W^{(l)}\bigg)$, where $\lambda_k(M), \sigma_k(M)$ represents the k-th eigenvalue and singular value, respectively, of the matrix $M$. Therefore, it appears clearly now that $\frac{1}{N_l} \text{Tr}\bigg(D^l W^{(l)})(D^l W^{(l)})^T\bigg)$ gives the empirical mean squared singular value of $D^l W^{(l)}$. 

Let us now show that $\lim_{N_l \to \infty} \frac{1}{N_l} \mathbb{E}_{W^{(l)}}\text{Tr}\bigg(D^l W^{(l)}(D^l W^{(l)})^T\bigg) = \chi $ in the infinite width limit and when $q^l$ is at its fixed point $q^*$.

\begin{align*}
\frac{1}{N_l} \mathbb{E}_{W^{(l)}}\text{Tr}\bigg(D^l W^{(l)}(D^l W^{(l)})^T\bigg) & = \frac{1}{N_l} \mathbb{E}_{W^{(l)}} \bigg(\sum\limits_{j=1}^{N_l} \sum_{i=1}^{N_{l-1}} \phi^\prime (h_i^{(l)})^2 (W_{ij}^{(l)})^2 \bigg) \\
& = \frac{1}{N_l} \sum\limits_{j=1}^{N_l} \sum_{i=1}^{N_{l-1}} \mathbb{E}_{W^{(l)}} \bigg(\phi^\prime (h_i^{(l)})^2 (W_{ij}^{(l)})^2 \bigg)\\
& = \frac{1}{N_l} \sum\limits_{j=1}^{N_l} \sum_{i=1}^{N_{l-1}} \phi^\prime (h_i^{(l)})^2  \mathbb{E}_{W^{(l)}} \bigg((W_{ij}^{(l)})^2 \bigg) \text{ considering $l$ big enough so that $q^l \approx q^*$} \\
& = \frac{1}{N_l} \sum\limits_{j=1}^{N_l} \sum_{i=1}^{N_{l-1}} \phi^\prime (h_i^{(l)})^2 \bigg( \gamma_l \frac{\sigma^2_{\alpha}}{N_{l-1}} \bigg) \\
& = \gamma_l \sigma_{\alpha}^2 \frac{1}{N_{l-1}} \sum_{i=1}^{N_{l-1}} \phi^\prime (h_i^{(l)})^2 \\
&\to \gamma_l \sigma^2_{\alpha} \int Dz \phi^\prime(\sqrt{q^*}z)^2 \text{ using the Law of Large Numbers with $N_{l-1} \to \infty$,}\\
& = \chi, 
\end{align*}
where we used, from the previous section, $\mathbb{E}_{W^{(l)}} \big((W_{ij}^{(l)})^2\big) = r_l \frac{\sigma^2_{\alpha}}{N_{l-1}} \frac{1}{N_l} = \gamma_l \frac{\sigma^2_{\alpha}}{N_{l-1}}$.

\subsection{Computation of $S_{W^TW}$ for low-rank Gaussian weights}\label{SM:Gaussian_weights}

Recall that $A_{ij} \overset{\text{iid}}{\sim}\mathcal{N}(0, \frac{\sigma^2_{\alpha}}{N})$ and $\text{rank}(A)=\gamma N$. Thus, its spectral density is given by the Marčenko Pastur distribution, where we first consider the matrix $\sigma_{\alpha}^{-2}A^TA$ as the variance of each coefficient is $\frac{1}{N}$ to make the computation simpler before appropriately rescaling the $\mathcal{S}$ Transform using the fact that if one rescales $B$ by $\sigma$, then $S_{\sigma B} = \sigma^{-1} S_{B}$.
\begin{align*}
\rho_{\sigma_{\alpha}^{-2} A^TA}(\lambda) & = (1 - \gamma)_+ \delta(\lambda) + \gamma \frac{\sqrt{(\lambda^+ - \lambda)(\lambda - \lambda^-)}}{2\pi\lambda} \mathds{1}_{[\lambda^-,\lambda^+]}(\lambda),\\
\intertext{where $x_+ = \text{max}(0,x), \lambda^- = (1-\frac{1}{\gamma})^2$ and $\lambda^+ = (1+\frac{1}{\gamma})^2$. The Stieltjes Transform is known to be }\\
G_{\sigma_{\alpha}^{-2} A^TA}(z) & = \gamma \frac{z + \gamma^{-1} - 1 - \sqrt{(\lambda^+ - z)(z - \lambda^-)}}{2z},\\
\intertext{from which we can easily compute the moment generating function}\\
M_{\sigma_{\alpha}^{-2} A^TA}(z) & = zG_{\sigma_{\alpha}^{-2} A^TA}(z) - 1 = \frac{1}{2} \big(-1 - \gamma + \gamma z - \gamma \sqrt{(\lambda^+ - z)(z - \lambda^-)} \big),\\
\intertext{whose invert is}
M^{-1}_{\sigma_{\alpha}^{-2} A^TA}(z) & = \frac{\gamma + z(1+\gamma) + z^2}{\gamma z}.\\
\intertext{And therefore}
S_{\sigma_{\alpha}^{-2} A^TA}(z) &= \frac{1+z}{zM^{-1}_{\sigma_{\alpha}^{-2} A^TA}(z) } = \gamma \frac{1+z}{\gamma + z(1+\gamma) + z^2} = \frac{1+z}{1 + z(1+\gamma^{-1}) + \gamma^{-1} z^2}.
\intertext{Note that when $\gamma=1$, the weight matrix is full rank and we get the same result as in \cite{Pennington_2018_emergence}. Rescaling the matrix by $\sigma^2_{\alpha}$ to match our original distribution gives}
S_{A^TA}(z) & = \sigma_{\alpha}^{-2} \frac{1+z}{1 + z(1+\gamma^{-1}) + \gamma^{-1} z^2}.\\
\intertext{Now note that as we have $W_{ij}\sim \mathcal{N}(0, \gamma \frac{\sigma^2_{\alpha}}{N})$, the scaling property of the $\mathcal{S}$ transform gives }
S_{W^TW} & = S_{(\sqrt{\gamma} A )^T \sqrt{\gamma} A} = \sqrt{\gamma}^{-2} S_{A^TA} = \gamma^{-1} S_{A^TA}.
\intertext{We can now expand $S_{W^TW}$ around 0 and identify from $S_{W^TW}(z):= \gamma^{-1} \sigma_{\alpha}^{-2} \big(1 + \sum\limits_{k=1}^\infty s_k z^{k}\big) $}
S_{W^TW}(z) & = \gamma^{-1}\sigma_{\alpha}^{-2} \frac{1+z}{1 + z(1+\gamma^{-1}) + \gamma^{-1} z^2} = \gamma^{-1} \sigma_{\alpha}^{-2} \big(1 - \frac{1}{\gamma} z + \frac{1-4\gamma}{\gamma^2}z^2 + \dots \big)\\
\implies s_1 & = -\frac{1}{\gamma}.
\end{align*}

\subsection{Computation of $S_{W^TW}$ for low-rank Orthogonal weights}\label{SM:orthogonal_weights}
Recall the spectral density of $W^TW$,
\begin{align*}
\rho_{\sigma_{\alpha}^{-2}{W}^TW}(z) & = \gamma \delta(z-1) + (1-\gamma) \delta(z).
\intertext{Then, the following computations are straightforward}
G_{\sigma_{\alpha}^{-2}{W}^TW}(z) & = \gamma (z-1)^{-1} + (1-\gamma) z^{-1},\\
M_{\sigma_{\alpha}^{-2}{W}^TW}(z) & = z G_{\sigma_{\alpha}^{-2}{W}^TW}(z) - 1 = \gamma (z-1)^{-1} \\
M_{\sigma_{\alpha}^{-2}{W}^TW}^{-1}(z) & = \frac{\gamma+z}{z} \\
S_{\sigma_{\alpha}^{-2}{W}^TW}(z) & = \frac{1+z}{zM_{\sigma_{\alpha}^{-2}{W}^TW}^{-1}(z)} = \frac{1+z}{\gamma + z} = \gamma^{-1}(1+z) (1+\gamma^{-1}z)^{-1} \\
\intertext{Rescaling by $\sigma_{\alpha}^2$, expanding around 0 and then identifying from $S_{W^TW}(z) := \gamma^{-1} \sigma_{\alpha}^{-2} \big(1 + \sum\limits_{k=1}^\infty s_k z^k\big)$  gives}
S_{{W}^TW}(z) & = \sigma_{\alpha}^{-2} S_{\sigma_{\alpha}^{-2}{W}^TW}(z) =  \sigma_{\alpha}^{-2} \gamma^{-1}(1+z) (1+\gamma^{-1}z)^{-1} \\
& = \sigma_{\alpha}^{-2} \gamma^{-1}(1+z) \sum\limits_{k=0}^\infty (-\frac{z}{\gamma})^k = \gamma^{-1} \sigma_{\alpha}^{-2} \big( 1 - (\gamma^{-1} - 1)z + (\gamma^{-2} - \gamma^{-1})z^2 + \dots \big)\\
\implies s_1 & = -(\gamma^{-1} - 1).
\end{align*}

%%%%%%%%%%%%%%%%%%%%%%%%%%%%%%%%%%%%%%%%%%%%%%%%%%%%%%%%%%%%%%%%%%%%%%%%%%%%%%%
%%%%%%%%%%%%%%%%%%%%%%%%%%%%%%%%%%%%%%%%%%%%%%%%%%%%%%%%%%%%%%%%%%%%%%%%%%%%%%%

\end{document}